\pdfoutput=1

\documentclass[11pt]{article}

\usepackage{acl}

\usepackage{times}
\usepackage[varqu,varl]{inconsolata}
\usepackage{latexsym}
\usepackage{booktabs}
\usepackage{siunitx}
\usepackage{nicefrac}

\usepackage[T1]{fontenc}
\usepackage{amsmath}
\usepackage{amsfonts}
\usepackage{graphicx}
\usepackage{soul}
\usepackage{overpic}
\usepackage[inline]{enumitem}
\usepackage{xspace}

\definecolor{pink}{RGB}{255, 71, 76}
\definecolor{kelleygreen}{RGB}{0, 147, 55}
\definecolor{purple}{RGB}{102, 102, 225}
\definecolor{lightyellow}{RGB}{255, 252, 187}

\sethlcolor{lightyellow}
\newcommand{\pl}[1]{\hl{#1}}

\usepackage[utf8]{inputenc}

\usepackage{microtype}

\title{Deduplicating Training Data Makes Language Models Better}

\author{
  Katherine Lee\thanks{\,\,\,\,Equal contribution. \textdagger\, Google Research, Brain Team. \ddag\,  University of Pennsylvania. Correspond to katherinelee@google.com and daphnei@seas.upenn.edu.}\,\,\textdagger\\
  \\\And
  Daphne Ippolito\footnotemark[1]\,\,\textdagger\ddag\\
  \\\And
  Andrew Nystrom\textdagger\\
  \\\And
  Chiyuan Zhang\textdagger\\
  \\\AND
  Douglas Eck\textdagger\\
  \\\And
  Chris Callison-Burch\ddag\\
  \\\And
  Nicholas Carlini\textdagger\\
}

\newcommand{\Approx}[0]{\textsc{NearDup}}
\newcommand{\Exact}[0]{\textsc{ExactSubstr}}
\newcommand{\Original}[0]{\textsc{Original}}

\begin{document}

\maketitle
\begin{abstract}
We find that existing language modeling datasets contain many near-duplicate examples and long repetitive substrings.
As a result, over $1\%$ of the unprompted output of language models trained on these datasets is copied verbatim from the training data.
We develop two tools that allow us to deduplicate training datasets---for example removing from C4 a single 61 word English sentence that is repeated over $60{,}000$ times.
Deduplication allows us to train models that emit memorized text ten times less frequently and require fewer training steps to achieve the same or better accuracy.
We can also reduce train-test overlap, which affects over $4\%$ of the validation set of standard datasets, thus allowing for more accurate evaluation.
Code for deduplication is released at
\url{https://github.com/google-research/deduplicate-text-datasets}.
\end{abstract}

\section{Introduction}

A key factor behind the recent progress in natural language processing is the development of large-scale text corpora used to train increasingly large language models.
These datasets have grown from single gigabytes to as much as a terabyte over the past few years \citep{chelba2013one,xue2020mt5,graff2003english,brown2020language}.
Because it is so expensive to perform manual review and curation on massive datasets, they tend to suffer in quality compared to their smaller predecessors.
This has implications far beyond metrics like perplexity and validation loss, as learned models reflect the biases present in their training data \cite{bender2021stochastic,wallace2019universal,sheng2020towards}.
Quantitatively and qualitatively understanding these datasets is therefore a research challenge in its own right \cite{dodge2021documenting}.


We show that one particular source of bias,
duplicated training examples, is pervasive:
all four common NLP datasets we studied contained duplicates. 
Additionally, all four corresponding validation sets contained text duplicated in the training set.
While naive deduplication is straightforward
(and the datasets we consider already perform some naive form
of deduplication), performing thorough deduplication at scale is both computationally challenging and requires sophisticated techniques.

We propose two scalable techniques to detect and remove duplicated training data.
\textit{Exact} substring matching identifies verbatim strings that are repeated.
    This allows us to identify cases where only part of a training example is duplicated (\S\ref{sec:exact}).
\textit{Approximate} full document matching uses hash-based techniques~\cite{broder1997resemblance} to identify pairs of documents with high $n$-gram overlap (\S\ref{sec:approx}).


We identify four distinct advantages to training on datasets that have been thoroughly deduplicated.
\begin{enumerate}

\item 
Over $1\%$ of tokens emitted unprompted from a model trained on standard datasets (e.g., C4) are part of a memorized sequence (See \S\ref{sec:memorization-results})---even though the 1.5 billion parameter model is much smaller than the 350GB dataset it was trained on.
By deduplicating the training dataset we reduce the rate of emitting memorized training data by a factor of $10\times$.

\item Train-test overlap is common in non-deduplicated datasets.
For example, we find \emph{a 61-word sequence}%
\footnote{``by combining fantastic ideas, interesting arrangements, and follow the current trends in the field of that make you more inspired and give artistic touches. We'd be honored if you can apply some or all of these design in your wedding. believe me, brilliant ideas would be perfect if it can be applied in real and make the people around you amazed!''} 
in C4 \citep{t52020} that is repeated $61{,}036$ times verbatim in the training dataset and $61$ times in the validation set ($0.02\%$ of the samples in each dataset).
This train-test set overlap not only causes researchers to over-estimate model accuracy, but also biases model selection towards models and hyperparameters that intentionally overfit their training datasets.

\item Training models on deduplicated datasets is more efficient.
Processing a dataset with our framework requires a CPU-only linear-time algorithm.
And so because 
these datasets are up to $19\%$ smaller, even including the deduplication runtime itself, training on deduplicated datasets directly reduces the training cost in terms of time, dollar, and the environment~\cite{bender2021stochastic, strubell2019energy, patterson2021carbon}.

\item Deduplicating training data does not hurt perplexity: models trained on deduplicated datasets have no worse perplexity compared to baseline models trained on the original datasets. 
In some cases deduplication reduces perplexity by up to $10\%$.
Further, because recent LMs are typically limited to training for just a few epochs \cite{radford2019language,t52020},
by training on higher quality data the models can reach higher accuracy faster.
\end{enumerate}
To summarize, data duplication offers significant advantages and no observed disadvantages.
In the remainder of this paper we present our text deduplication framework in \S\ref{sec:methods}, and study the extent of duplicate content in common NLP datasets (e.g., C4, Wiki-40B, and LM1B) in \S\ref{sec:deduplication-results}.
We then examine the impact of deduplication on test perplexity (\S\ref{sec:perplexity-results}) and on the frequency of emitting memorized content (\S\ref{sec:memorization-results}).
Finally, we analyze to what extent perplexity on existing, released models are skewed as a result of overlap between the train and test/validation splits (\S\ref{sec:eval-existing-models}).

\section{Related Work}
\paragraph{Large language model datasets.}
While we believe our results are independent of model architecture,
we perform our analysis on Transformer-based decoder-only language models \citep{vaswani2017attention} trained for open-ended text generation.
These current state-of-the-art models are trained on internet text.
For example, the GPT-2 family of models \citet{radford2019language} is trained on WebText, a dataset of web documents highly ranked on Reddit---however this dataset was not made available publicly.
A common dataset starting point is CommonCrawl, an index of public webpages.
Among the models trained on CommonCrawl include
GPT-3 \cite{brown2020language} with the addition of book datasets,
GROVER \cite{zellers2019defending} on a restricted subset filtered to news domains called RealNews,
and T5 \cite{t52020} on a cleaned version of common crawl called C4.
Other models are trained on more curated Internet sources---for example \citet{guo2020wiki40b} used high quality processed Wikipedia text from 40 different languages to train monolingual 141.4M parameter language models.
Non-English models necessarily use different datasets; \citet{zeng2021pangualpha} for instance introduced PANGU-$\alpha$, a family of models with up to 200B parameters that were trained on a non-public corpus of cleaned and filtered Chinese-language documents from CommonCrawl and other sources.
Since many of these datasets are not public,
we deduplicate three that are: Wiki-40B, C4, and RealNews--as well as the One Billion Word Language Model Benchmark \citep{chelba2013one}, 
a smaller
dataset commonly used for evaluation.

\paragraph{Contamination of downstream tasks.}
When models are trained on datasets constructed by crawling the Internet, it is possible the model will train on the test set of downstream target tasks.
For example, \citet[\S{}4]{radford2019language} performed a post-hoc analysis to identify 8-gram overlaps between GPT-2's training set and datasets used for evaluation,
and \citet{Dodge2021-lb} analyzed C4 and found that up to 14.4\%  of test examples for various standard tasks were found verbatim (normalizing for capitalization and punctuation) in the dataset.
A more proactive approach removes contaminated data.
\citet[Appendix B]{trinh2018simple} removed documents from their CommonCrawl-based train set that overlapped substantially with the commonsense reasoning used for evaluation.
And GPT-3 \cite[\S{}5]{brown2020language} did the reverse and removed downstream evaluation examples from their training data by conservatively filtering out any train set examples with a 13-gram overlap with any evaluation example.
Up to $90\%$ of tasks were flagged as potentially contaminated.

In our research, we do not focus on the impact of duplicate text in pretrained models on downstream benchmark tasks; instead we address how duplicate text in the LM training and validation sets impacts model perplexity and the extent to which generated text included memorized content.

\paragraph{Memorizing training data.} The privacy risks of data memorization, for example the ability to extract sensitive data such as valid phone numbers and IRC usernames, are highlighted by
\citet{carlini2020extracting}.
%
While their paper finds 604 samples that GPT-2 emitted from its training set, we show that \emph{over $1\%$} of the data most models emit is memorized training data.
In computer vision, memorization of training data has been studied from various angles for both discriminative and generative models~\citep[e.g.][]{arpit2017closer,8953411,feldman2020neural,stephenson2021geometry,teterwak2021understanding}.

\paragraph{Duplicate text in training data.}
The Book Corpus \citep{zhu2015aligning}, which was used to train popular models such as BERT, has a substantial amount of exact-duplicate documents according to \citet{bandy2021addressing}.
\citet{allamanis2019adverse} shows that duplicate examples in code datasets cause worsened performance on code understanding tasks.


\section{Language Modeling Datasets}
We analyze the presence of duplicate text in four datasets of varying sizes that have been used for training natural language generation systems, producing general-purpose pre-trained models, and for language model benchmarking.
While this paper restricts itself to English datasets, we expect that non-English datasets suffer from similar issues and could likewise benefit from de-duplication.

\paragraph{Wikipedia (Wiki-40B)}
consists of multi-lingual cleaned Wikipedia text \citep{guo2020wiki40b}.
We take the English portion, which contains 2.9M Wikipedia pages with an average length of 768 BPE tokens.
The dataset creators do not indicate any deduplication was performed aside from removing redirect-pages (e.g., ``sunflower'' to ``Helianthus'').

\paragraph{One-Billion Word benchmark (LM1B)}  contains 30M sentences of news commentary \citep{chelba2013one}.
Unlike the other datasets we analyze, LM1B's examples are one sentence long rather than multi-sentence documents.
The average example length is 32 BPE tokens.
While this dataset is extremely standard for benchmarking language models, \citet[Sec 4]{radford2019language} note it has 13.2\% overlap of the test set with the train set.

\paragraph{Colossal Cleaned Common Crawl (C4)}
is made up of 360M web documents, with an average length of 486 BPE tokens \citep{t52020}.
C4 was introduced as a pre-training dataset for T5, a set of encoder-decoder models which have been widely used in fine-tuned downstream tasks.
The dataset was previously deduplicated in a more sophisticated process
than the prior two datasets.
Each paragraph was hashed and paragraphs resulting in hash collisions were removed.
This was followed by a pass that removed placeholder text, code, and prohibited words.
See \citet{dodge2021documenting} for a detailed breakdown of the source text in C4.

\paragraph{RealNews}
is a subset of the Common Crawl consisting of articles from news domains \citep{zellers2019defending}.
It contains 31M documents with average length 793 BPE tokens.
RealNews was deduplicated by inserting a hash of the first 100 characters of each document into a bloom filter \citep{bloom1970space} and then excluding any document which resulted in a hash collision.
Like C4, examples with duplicate URLs were excluded.

\section{Methods for Identifying Duplicates}
\label{sec:methods}

The simplest technique to find duplicate examples would be to perform exact string matching between all example pairs, but as we will show, this is insufficient.
We introduce two complementary methods for performing deduplication.
First, using a suffix array \cite{manber1993suffix}, we remove duplicate substrings from the dataset if they occur verbatim in more than one example.
Second, we use MinHash \citep{broder1997resemblance}, an efficient algorithm for estimating the $n$-gram similarity between all pairs of examples in a corpus, to remove entire examples from the dataset if they have high $n$-gram overlap with any other example.

We consider a dataset $D = \{x_i\}_{i=1}^N$ as a collection of \emph{examples} $x_i$.
Each of these examples is itself a sequence of \emph{tokens}: $x_i = \left[ x_i^1, x_i^2, \cdots, x_i^{s_i} \right]$.

\subsection{Exact Substring Duplication} \label{sec:exact}
Due to the diversity of possibilities in human language, it is rare for the same idea to be expressed identically in multiple documents unless one expression is derived from the other, or both are quoting from a shared source.
This observation motivates deduplicating exact substrings. We call our approach \Exact{}.
When two examples $x_i$ and $x_j$ share a sufficiently long substring (that is, a substring for which $x_i^{a..a+k} = x_j^{b..b+k}$), that substring is removed from one of them.
Based on statistical analyses (\S\ref{section:exact_thresh}), we select $k=50$ tokens as the minimum matching substring length.
A breakdown of the computation needed for this approach can be found in Appendix \ref{sec:suffix-implementation}.

\subsubsection{Suffix Arrays}
This exact-substring-matching criterion, while conceptually simple, is computationally prohibitive with naive (quadratic) all-pair matching.
To improve the efficiency, we concatenate all the examples of the entire dataset $D$ into a giant sequence $\mathcal{S}$, and construct a Suffix Array $\mathcal{A}$ of $\mathcal{S}$.
A suffix array \citep{manber1993suffix} is a representation of a suffix tree \citep{weiner1973linear} that can be constructed in linear time in $\lVert \mathcal{S} \rVert$ \citep{karkkainen2003simple}  
and enables efficient computation of many substring queries; in particular, they allow us to identify duplicated training examples in linear time.
Suffix arrays have the advantage over suffix trees in that they are 10--100$\times$
more memory efficient \cite{manber1993suffix}, requiring just 8 bytes per input token, though they are asymptotically less
efficient for some query types.
They have been used widely in NLP, such as for efficient TF-IDF computation \citep{yamamoto2001using} and document clustering \citep{hung2007new}.

The suffix array $\mathcal{A}$ for a sequence $\mathcal{S}$ is a lexicographically-ordered list of all suffixes contained in the sequence. 
%
Formally,
\[ \mathcal{A}(\mathcal{S}) = \mathop{\text{arg sort}} \text{all\_suffixes}(\mathcal{S}) \]
For example, the suffixes of the sequence ``banana'' are (``banana'',  ``anana'', ``nana'' ``ana'', ``na'', ``a'')
and so the suffix array is the sequence (6 4 2 1 5 3).
In practice, we construct $\mathcal{S}$ from the bytes of the BPE tokenization of the text (\S\ref{sec:impact-trained-models}).



\subsubsection{Substring matching}

After constructing $\mathcal{A}$, it is straightforward to identify duplicated training examples.
Suppose that the sequence $s$ was repeated exactly twice in the training dataset $\mathcal{S}$ at positions $i$ and $j$,
that is, $\mathcal{S}_{i..i+|s|} = \mathcal{S}_{j..j+|s|}$.
Then the indices $i, j$ will occur adjacent to each other in the suffix array $\mathcal{A}$.

Finding all repeated sequences is thus a matter of linearly scanning the suffix array from
beginning to end and looking for sequences $\mathcal{A}_i, \mathcal{A}_{i+1}$ that share a common prefix of
at least some threshold length.
Any satisfying sequences are recorded.
This algorithm is embarrassingly parallel, and so we can efficiently process the dataset.
Based on experimentation (Appendix \ref{section:exact_thresh}), we choose a threshold length of 50 BPE tokens for all experiments.

\begin{table*}[htbp]
  \scriptsize
  \centering
    \begin{tabular}{l|p{0.39\linewidth}|p{0.41\linewidth}}
    \toprule
    \multicolumn{1}{c|}{Dataset} & \multicolumn{1}{c|}{Example} & \multicolumn{1}{c}{Near-Duplicate Example} \\
    \midrule
    Wiki-40B & \pl{\textbackslash{}n\_START\_ARTICLE\_\textbackslash{}nHum Award for } {Most Impactful Character} \pl{\textbackslash{}n\_START\_SECTION\_\textbackslash{}nWinners and nominees\textbackslash{}n\_START\_PARAGRAPH\_\textbackslash{}nIn the list below, winners are listed first in the colored row, followed by the other nominees.} [...] &
    \pl{\textbackslash{}n\_START\_ARTICLE\_\textbackslash{}nHum Award for} {Best Actor in a Negative Role} \pl{\textbackslash{}n\_START\_SECTION\_\textbackslash{}nWinners and nominees\textbackslash{}n\_START\_PARAGRAPH\_\textbackslash{}nIn the list below, winners are listed first in the colored row, followed by the other nominees.} [...] \\
    \midrule
    LM1B  & \pl{I left for California in 1979 and tracked Cleveland 's changes on trips back to visit my sisters .} & \pl{I left for California in 1979} , \pl{and tracked Cleveland 's changes on trips back to visit my sisters .} \\
    \midrule
    C4    & \pl{Affordable and convenient holiday flights take off from your departure country,} "Canada"\pl{. From} May \pl{2019 to October 2019, Condor flights to your dream destination will be roughly} 6 \pl{a week! Book your} Halifax (YHZ) - Basel (BSL) \pl{flight now, and look forward to your} "Switzerland" \pl{destination!} &
    \pl{Affordable and convenient holiday flights take off from your departure country,} "USA"\pl{. From} April \pl{2019 to October 2019, Condor flights to your dream destination will be roughly} 7 \pl{a week! Book your} Maui Kahului (OGG) - Dubrovnik (DBV) \pl{flight now, and look forward to your} "Croatia" \pl{destination!} \\
    \bottomrule
    \end{tabular}%
  \caption{Qualitative examples of near-duplicates identified by \Approx{} from each dataset. The similarity between documents is highlighted. Note the small interspersed differences that make exact duplicate matching less effective. Examples ending with ``[...]'' have been truncated for brevity.
  More data available in Appendix.}
  \label{tab:qualitative_examples}%
\end{table*}%

\subsection{Approximate Matching with MinHash} \label{sec:approx}

We also perform \emph{approximate} deduplication based on matching entire examples.
This method, which we call \Approx, is a good complement to the \emph{exact} substring matching, especially for web crawl text, as it handles the very common case of documents being identical except for interspersed templated fields (such as the last row of Table \ref{tab:qualitative_examples}).

MinHash \citep{broder1997resemblance} is an approximate matching algorithm widely used in large-scale deduplication tasks \citep{versley2012not,GABRIEL201863,gyawali2020deduplication}, including to deduplicate the training set for a large Chinese-language LM \citep{zeng2021pangualpha}.
Given two documents $x_i$ and $x_j$, the main idea is to represent each document by its respective set of $n$-grams $d_i$ and $d_j$.
We can then use hash functions to approximate the \emph{Jaccard Index} \citep{jaccard1912distribution}:
\begin{equation*}
\operatorname{Jaccard}(d_i, d_j) = \nicefrac{|d_i \cap d_j|}{|d_i \cup d_j|}
\end{equation*}
If the Jaccard Index between $d_i$ and $d_j$ is sufficiently high, it is likely that documents are approximate matches of each other.
To efficiently approximate the Jaccard index, MinHash constructs document signatures by sorting each of the $n$-grams via a hash function, and then keeping only the $k$ smallest hashed $n$-grams.
There are multiple ways to construct estimators of the Jaccard index from these kinds of signatures \citep{cohen2016min}.

In our implementation, we use 5-grams and a signature of size 9,000. The probability that two documents are considered a potential match is
\begin{equation*}
\operatorname{Pr}(d_i, d_j | \operatorname{Jaccard}(d_i, d_j) = s_{i, j}) = 1 - (1 - s_{i, j}^b)^r
\end{equation*}
where $b=20$ and $r=450$ are user-settable parameters to control the strength of the filter.
See Appendix~\ref{section:minhash_details} for more details.

For each pair of documents identified as a potential match, more computationally expensive similarity metrics can be employed as a subsequent filtering step.
In particular, we identify two documents as duplicates if they are matched by the MinHash algorithm and their \emph{edit similarity} is greater than 0.8. The edit similarity between token sequences $x_i$ and $x_j$ is defined as:
\begin{equation*}
    \operatorname{EditSim}(x_i, x_j) = 1 - \frac{\operatorname{EditDistance}(x_i, x_j)}{\max(|x_i|, |x_j|)}
\end{equation*}

\noindent To build clusters of similar documents, we construct a graph that has an edge between two documents if they are considered a match. Then, we use the method introduced in \citet{lacki2018connected} to identify  connected components.
A breakdown of the computation needed is given in Appendix \ref{section:minhash_details}.

\begin{figure}[t]
    \centering
    \includegraphics[width=\linewidth]{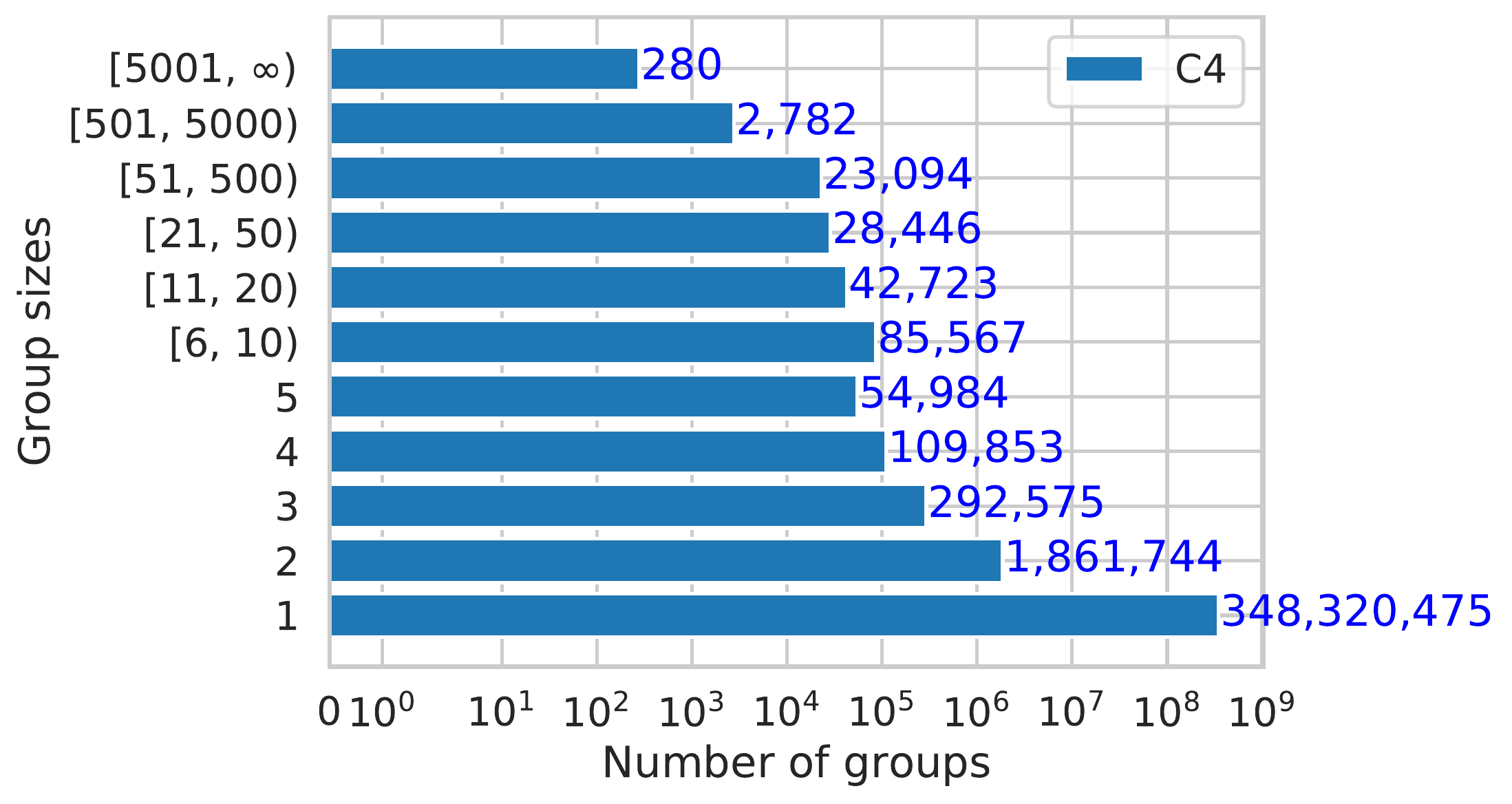}
    \caption{The distribution of near-duplicate cluster sizes from running \Approx{} on C4.}
    \label{fig:nd3-cluster-hist-c4}
\end{figure}

\section{Deduplication Results}\label{sec:deduplication-results}
We deduplicate each of the four datasets with both of our two techniques.
When text was duplicated across multiple data splits, we prioritized keeping a copy in the test or validation set and removing it from the train set.
%


\subsection{Amount of Text Removed}
With \Approx{}, we found that the web-scrape datasets contain between 3.04\% (on C4) to 13.63\% (on RealNews) near duplicates (Table \ref{tab:num_duplicates}).
Near-duplicate text is much less common in Wiki-40B, forming only 0.39\% of the train set.\footnote{Most duplicates we saw were automatically generated pages, such as the outcomes of sports games.
This shows the strength of manual curation for creating high-quality datasets.}
In C4, the majority (1.8M) of near-duplicate clusters consisted of just a single pair of examples that matched against each other, but there were 280 clusters with over 5,000 examples in them (Figure \ref{fig:nd3-cluster-hist-c4}), including one cluster of size 250,933.

On average with \Exact{}, we remove more total content than with \Approx{} (despite \Exact{} not removing any examples outright)---for example removing $7.18\%$ of the tokens in C4.
The exception is LM1B, where \Exact{} removes $8\times$ less data than
\Approx{}.
On investigation, we find this is due to the fact that LM1B documents are significantly shorter: $90\%$ of all documents are under 50 tokens, and so are not even candidates for potential matches even if the entire sequence matched verbatim.
We find that both \Approx{} and \Exact{} remove similar content---$77\%$ of the training examples that \Approx{} removes from C4 have at least one verbatim length-$50$ match found by \Exact{}.


    
\begin{table}[tbp]
  \centering
  \small
    \begin{tabular}{l|rr|r}
    \toprule
          & \multicolumn{2}{c}{\% train examples with} & \multicolumn{1}{c}{\% valid with} \\
          & \multicolumn{1}{c}{dup in train} & \multicolumn{1}{c}{dup in valid} & \multicolumn{1}{c}{dup in train} \\
          \midrule
    C4    & 3.04\% & 1.59\% & 4.60\%  \\
    RealNews & 13.63\% & 1.25\% & 14.35\%  \\
    LM1B  & 4.86\% & 0.07\% & 4.92\%  \\
    Wiki40B & 0.39\% & 0.26\% & 0.72\% \\
    \bottomrule
    \end{tabular}%
  \caption{The fraction of examples identified by \Approx{} as near-duplicates.}
  \label{tab:num_duplicates}%
\end{table}%

\begin{table}[tbp]
  \centering
  \small
    \begin{tabular}{l|S[table-format=3.2]S[table-format=3.3]|S[table-format=3.3]}
    \toprule
          & \multicolumn{2}{c}{\% train tokens with} & \multicolumn{1}{c}{\% valid with} \\
          & \multicolumn{1}{c}{dup in train} & \multicolumn{1}{c}{dup in valid} & \multicolumn{1}{c}{dup in train} \\
          \midrule
    C4    & 7.18\% & 0.75\%  & 1.38\% \\
    RealNews & 19.4\%  & 2.61\%  & 3.37\% \\
    LM1B  & 0.76\%  & 0.016\%  & 0.019\% \\
    Wiki40B & 2.76\%  & 0.52\%  & 0.67\% \\
    \bottomrule
    \end{tabular}%
  \caption{The fraction of tokens  (note Table~\ref{tab:num_duplicates} reports the fraction of \emph{examples}) identified by \Exact{} as part of an exact duplicate 50-token substring.}
  \label{tab:num_duplicates2}%
\end{table}%

\subsection{Properties of Duplicated Text}
While the authors of both RealNews and C4 explicitly attempted deduplication during dataset construction, the methods were insufficient to capture the more subtle types of duplicate text commonly found on the internet.
In C4 and Wiki-40B, we qualitatively observe that much of the text identified as near-duplicated is computer-generated.
The text is identical except for the names of places, businesses, products, dates, and so on. 
Because these examples frequently differ by just a few words at a time, deduplication strategies relying on exact string matching would fail to identify a match.
Example duplicate pairs from each dataset can be found in Table \ref{tab:qualitative_examples}  (more examples in the Appendix).

For RealNews and LM1B, derived from news sites, we observe that many near-duplicates occur because the same news article appears on multiple news sites with slightly different formatting.
For example, in LM1B, there is one example that starts ``\textit{MINEOLA , N.Y. - New York officials say} [...]'' and another that starts ``\textit{( AP ) - New York officials say} [...]''.
The two examples are otherwise identical.

\subsection{Train / Test Set Leakage}
\label{sec:leakage}
Both deduplication methods identify overlap between the train set and the validation set (Table \ref{tab:num_duplicates}).
For example, 4.6\% of the C4 validation set and 14.4\% of the RealNews validation set examples had an approximate duplicate in their respective training sets.
Such duplication is problematic since it could cause evaluation metrics to be unfairly inflated for models that are better at memorizing their train sets.
We evaluate the effect of this leakage on publicly released models in Section \ref{sec:eval-existing-models}.

\section{Impact on Trained Models}
\label{sec:impact-trained-models}.
We trained 1.5B parameter ``XL", decoder-only, Transformer-based language models similar to GPT-2, on C4-\Original, C4-\Approx, and C4-\Exact, respectively.
We use the T5 codebase and model architecture from \citet{t52020}, and each model was trained for about two epochs on its respective dataset.
To better understand the amount of variance in the perplexities of trained models, we also trained three different random seeds of the 110M parameter ``base" model for each of the above three datasets---for a total of nine base-sized models.

For all experiments, we used a Byte Pair Encoding (BPE) vocabulary trained on C4-\Approx{} with a budget of 50K tokens, which resulted in a vocabulary the same size as GPT-2's.
We trained with a maximum sequence length of 512 tokens (for longer documents, we randomly extracted subsequences of this length.)
Further training details can be found in Appendix \ref{sec:model-training-details}.

\subsection{Model Perplexity}\label{sec:perplexity-results}

\begin{figure}[t]
    \centering
    \includegraphics[width=\linewidth]{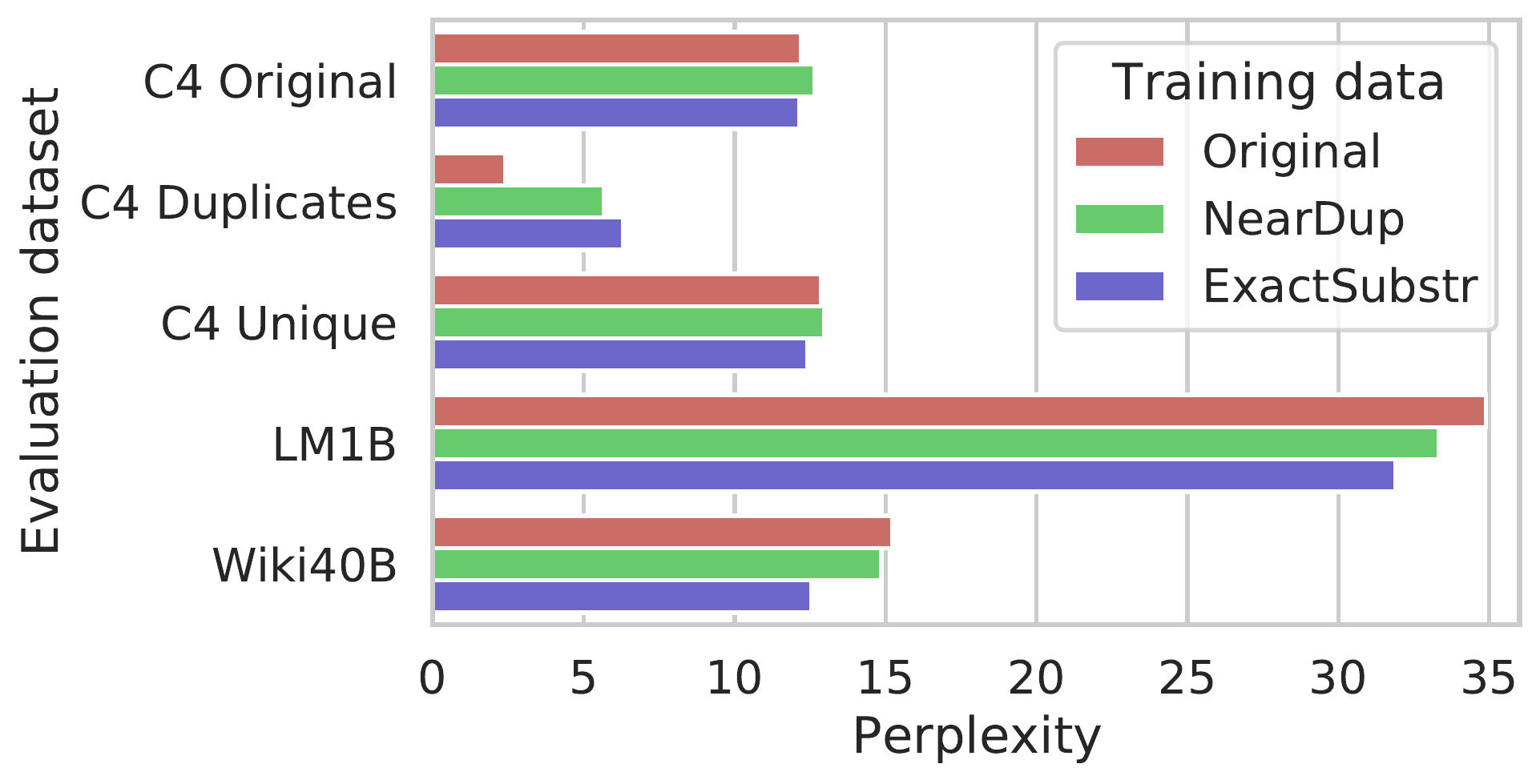}
    \caption{
Impact of deduplicating the training set on validation perplexity.
We plot the results from T5 XL (see Appendix for base-sized model).
For C4, we evaluate on \textit{C4 Original}, the original validation set; \textit{C4 Unique}, a subset of the validation set identified by \Approx{} as having zero matches across C4; and \textit{C4 Duplicates}, a subset of the validation set identified by \Approx{} as having a match in the C4 train set.
}
\label{fig:eval-ppl}
\end{figure}

We computed the perplexity of our trained models on the validation sets of LM1B and Wiki-40B, and on subsets of the C4 validation set (Figure \ref{fig:eval-ppl}).
For the base size, we observe that all models have similar perplexity on the original C4 validation set and on validation set examples that were identified as unique (no near-duplicate in either train or validation).
However, both models trained on deduplicated data have significantly higher perplexity on validation set examples that have duplicates in the training set than the model trained on the original C4. \Exact-deduplicated results in higher perplexity than \Approx-deduplicated.
These trends holds true for the XL sized model as well.
While this may suggest \Exact{} duplication results in models least overfit on the train set, note that both of these techniques have
used separate duplicate thresholds and a different choice of thresholds could change the results.

When evaluating on the validation sets of LM1B and Wiki-40B, we found that models trained on \Approx-deduplicated C4 consistently achieved lowest perplexity (for LM1B eval with base models, see Appendix Figure \ref{fig:eval-ppl-with-lm1b}).
\Exact{} deduplication decreases perplexity of the XL model by almost 3 points perplexity on Wiki-40B which is much larger than the variation of about 1 point perplexity we observed in the base models.
This is despite seeing fewer tokens of training data overall.

Lastly, we note all our XL models achieved  <35 perplexity on LM1B, which is less than the 42.16 perplexity reported for the 1.5B GPT-2 using a vocabulary the same size as ours.

\subsection{Generated Text}\label{sec:memorization-results}
Data duplication has the effect of biasing the trained LM towards particular types of examples. 
This can contribute to a lower diversity of generations, and increased likelihood that the generated content is copied from the training data \citep{carlini2020extracting}.
For our generation experiments, we use top-$k$ random sampling with $k=50$ and experiment with prompted and unprompted generation.

\begin{table}[t]
    \centering
    \small
\begin{tabular}{l|rr}
\toprule
Model & 1 Epoch & 2 Epochs \\
\midrule
XL-\Original & 1.926\% & 1.571\% \\
XL-\Approx & 0.189\% & 0.264\% \\
XL-\Exact & 0.138\% & 0.168\% \\
\bottomrule
\end{tabular}
\caption{When generating 100k sequences with no prompting, over $1\%$ of the tokens emitted from a model trained on the original dataset are part of a 50-token long sequence copied directly from the training dataset. This drops to $0.1\%$ for the deduplicated datasets.}
\label{tab:memorizations_no_prompt}
\end{table}
\paragraph{No prompt.}
We first evaluate memorization tendencies in the case where the model is asked to generate text without any prompt sequence.
We generate 100,000 samples, each up to 512 tokens in length (examples provided in the Appendix).
For each generated token, we say the token is memorized if it is part of a 50-token substring that is exactly contained in the training data.
On XL-\Original, over 1\% of the generated tokens belong to memorized sub-sequences (see Table~\ref{tab:memorizations_no_prompt}).
This is $\sim10\times$ more memorization than XL-\Exact{} or XL-\Approx.
Some example subsequences that were copied verbatim from the train set can be found in Table \ref{tab:exact_substr_examples} in the Appendix.

\begin{figure}
    \centering
    \small
    \includegraphics[width=\linewidth]{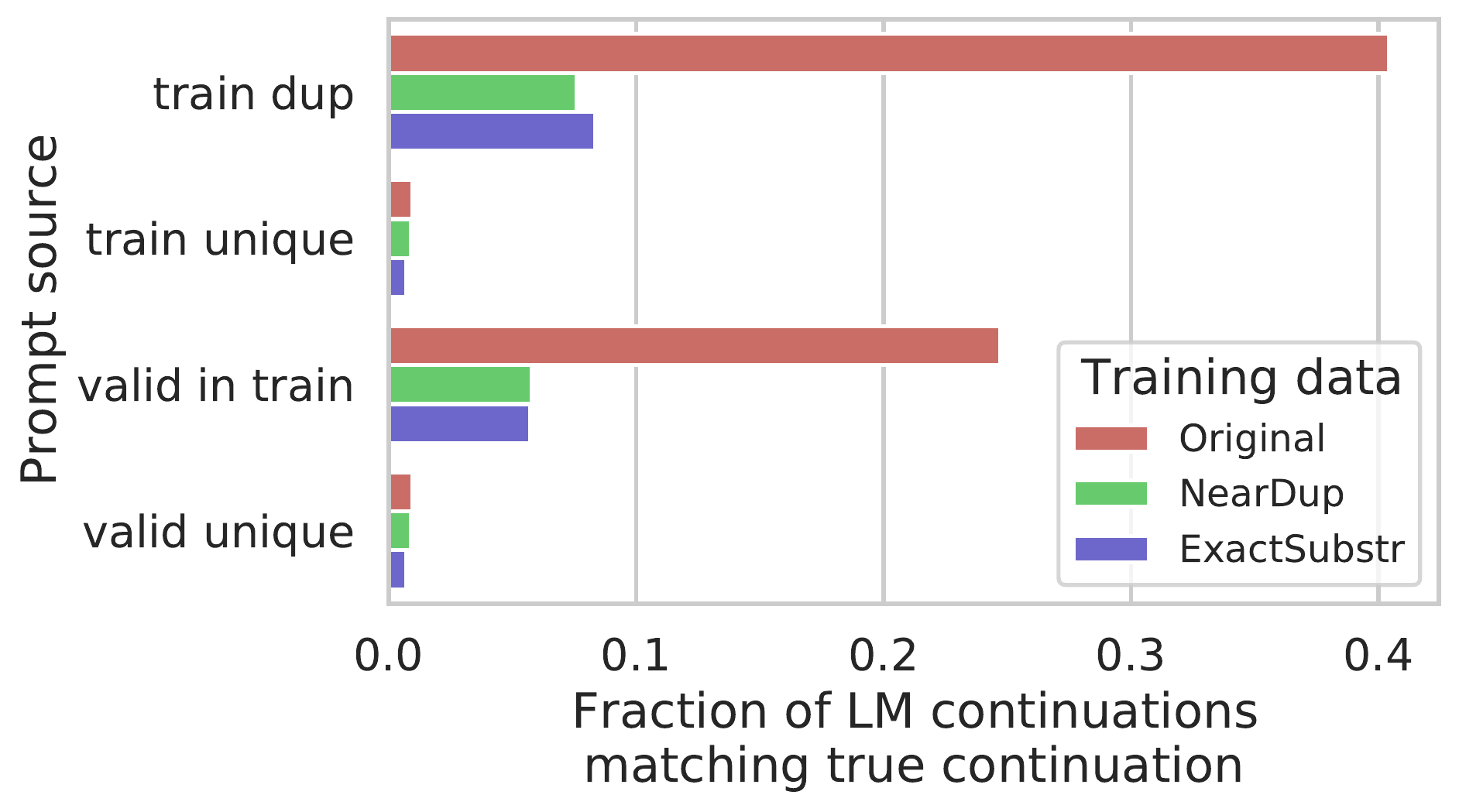}
    \caption{The proportion of generations which have edit similarity above 0.8 with the groundtruth continuation when using the LM to generate continuations for 32-token prompts identified by \Approx{} as either duplicated or unique.}
    \label{fig:ground-truth-continuation}
\end{figure}
\paragraph{With prompting.}
In most real use cases, language model generation is controlled by providing a prompt for the model to continue.
We experiment with four possible prompt sources: training examples identified by \Exact{} as having near-duplicates in the train set (train dup), training examples identified as unique (train unique), validation set examples with a near-duplicate in the train set (valid in train), and validation set examples identified as unique across all splits (valid unique).
We select the first 32 tokens of each example as the prompt, which means we can evaluate the fraction of generations which are near-duplicates with the ground-truth continuation for the prompt (Figure \ref{fig:ground-truth-continuation}).
When the prompt comes from duplicate examples in the train set, XL-\Original{} reproduces the groundtruth continuation over 40\% of the time.
XL-\Exact{} and XL-\Approx{} still copy the groundtruth more often when the prompt comes from a duplicate example than when the prompt comes from a unique example, suggesting that more stringent deduplication may be necessary to remove memorization tendencies entirely. 


\begin{table}[]
    \centering
    \small
    \begin{tabular}{ll|rrr}
    \toprule
    Model & Dataset & \multicolumn{1}{c}{Orig} & \multicolumn{1}{c}{Dups} & \multicolumn{1}{c}{Unique} \\
    \midrule
    Transformer-XL & LM1B & 21.77 & 10.11  & 23.58  \\
    GROVER-Base  & RealNews & 15.44 & 13.77 & 15.73  \\
    GROVER-XL & RealNews & 9.15 & 7.68 & 9.45 \\
    \bottomrule
    \end{tabular}
    \caption{For each model, the perplexity of the official validation set (\textit{Orig}), valid set examples which were identified by \Approx{} as matches of train set examples (\textit{Dups}), and valid set examples identified by \Approx{} as unique (\textit{Unique}).
    Due to the size of the RealNews validation set, we evaluated on only the first 25k examples meeting each condition.}
    \label{tab:ppl_sota_models}
\end{table}
\subsection{Impact on Existing Models} \label{sec:eval-existing-models}
Train-test leakage does not just impact models trained on C4.
Table \ref{tab:ppl_sota_models} shows that
the presence of near-duplicates of the evaluation set
in the train set has a significant impact on model perplexity for two standard models: Transformer-XL \citep{dai2019transformer}, which was trained on LM1B, and GROVER \citep{zellers2019defending}, which was trained on RealNews.
For Transformer XL, the perplexity halves on examples identified as near-duplicates.
For GROVER, the difference, though not quite as stark, is present in both model sizes considered.

Existing models also suffer from the problem of generating text from their train sets.
We find that $1.38\%$ of the tokens in the official release of 25k GROVER-Mega outputs
\footnote{\url{gs://grover-models/generation_examples/generator=mega~dataset=p0.90.jsonl}}
are part of verbatim matches in RealNews of at least length $50$.
Likewise, more than 5\% of the tokens in \textasciitilde 200k sequences outputted by GPT-Neo 1.3B \citep{gpt-neo} are part of a $50$ token matches of its training data, the Pile \citep{pile}.

\section{Discussion}
The focus of this paper is on the datasets used to train language models.
While recent work focused on documenting the potential harms that could arise from problematic datasets  \cite{bender2018data, gebru2020datasheets}, less work has been done to 
quantitatively analyze properties of real language modelling datasets, like \citet{dodge2021documenting} has done for C4.
Our paper provides analysis on one particular axis, that of data duplication.

Our experiments measured what could be quantified: the amount of duplicate content in common datasets, the effect of deduplication on trained model perplexity, and the reduction of memorized content in trained models through deduplication.
We do not focus on the nature of the data being removed by deduplication or memorized by LMs.

Privacy is an important subject for future work, as memorized training data has significant privacy consequences.
By this, we mean the standard privacy definition that a model should not reveal anything particular to the specific dataset it was trained on, as opposed to another training dataset from a similar distribution \citep{shokri2017membership}.\footnote{%
Another interpretation of privacy focuses on the sensitivity of the data involved, when a model is trained on and able to reproduce personal identifiers or other forms of ``private data.'' Our definition is more expansive.}
Training on standard datasets that have not yet been deduplicated results in models that are particularly sensitive to examples that happened to be repeated multiple times, and this has negative privacy implications.
For instance, it could violate a person's expectations of privacy if their publicly available personal data appeared in a different, surprising context.
Downstream applications of LMs, such as the game AI Dungeon\footnote{\url{https://play.aidungeon.io/}}, should also not output memorized content like adverts for real products. 

We stress that in our experiments, we do not distinguish between undesired memorized text (such as phone numbers), innocuous memorized text (common phrases), and text we may want to be memorized (such as a quote by a public figure), and instead treat all instances of the LM generating text that closely matches the training set as problematic.
While we qualitatively observed that much of the identified memorized content was relatively innocuous, a more systematic study of the risks associated with the detected memorization was beyond the scope of this work.

We also do not investigate the negative consequences of deduplication.
Some language tasks explicitly require memorization, like document retrieval or closed-book question answering. 
Also, text that gives attribution is often duplicated across documents, so
removing duplicate substrings could correspond to removing \emph{just} the attribution, which could result in models that learn the content without its attached attribution.
Deduplication is also not sufficient to remove privacy-sensitive data like bank passwords and medical records which should never be used in training data \cite{brown2022privacy}.

Ultimately, whether memorization is a desired property of a language model, or else risky and unwanted, depends
both on the nature of the text that has been memorized and on the downstream applications of the trained model.
However, since the trend has been towards creating datasets and models that are application-agnostic, we encourage researchers to think carefully about the limitations of the data they have
collected and the how the model's intended usage constrains what should be part of the training set. 
Developing techniques to memorize or forget specific sequences depending on the end application is a promising research direction. 

\section{Conclusion}

We encourage future language model research to perform dataset deduplication, either by training on the deduplicated datasets we release, using the deduplication tools we release, or following our approach to deduplicate datasets with new tools.

The exact technique used to perform deduplication is less important than performing stringent deduplication in the first place.
On the whole, deduplication does not harm, and sometimes improves, model perplexity, despite the fact that the deduplicated datasets are smaller and faster to train on.
It is especially important that there are no duplicates between the training and testing sets, because overlap here explicitly encourages selecting models that memorize the training data.
Lastly, deduplication helps to reduce some of the privacy concerns around LMs memorizing their training data.

\section*{Ethics}

The developers of large language models typically attempt to create training data that reflects natural human communication, but current methods to collect and curate such datasets are fallible.
There are multiple reasons some text ends up over-represented. 
For example, bot replies, auto-generated templates, and licenses are repeated for structural (e.g., legal, economical) reasons (as was also observed by \citet{dodge2021documenting}).
Additionally, common techniques for acquiring and ``cleaning'' data can result in an over-representation of particular subsets of world users, often those who are English-speaking and publishing in established forums.
This effectively under-represents non-English speakers as well as groups whose communication mostly occurs outside of the public web. 
In this paper, we focus on the problem of over-representation of some types of text (structural duplicates) but do not address the problem of under-representation of others.

Additionally, while we discuss when memorized content might be desired and when it might not be desired, our analysis does not disambiguate these two cases.
Work to disambiguate helpful from harmful memorization is tremendously complex and would require a different set of research methodologies than are presented in this work. 

\section*{Acknowledgements}
We are grateful to the many researchers whose technical help, feedback, and discussions shaped this project:
Jacob Austin,
Samy Bengio,
Olivier Bousquet,
James Bradbury,
Fernando Diaz,
Mark Diaz,
Noah Fiedel,
Jonathan Frankle,
David Grangier,
Stefanie Karp,
David Mimno,
Gaurav Mishra,
Michael Mozer,
Sharan Narang,
Alex Passos,
Adam Roberts,
Hanie Sedghi,
Jascha Sohl-dickstein,
David So,
Florian Tramer,
and 
Yun William Yu.
We are also grateful to the Google Brain women who have given us continuous support.

Chris Callison-Burch and Daphne Ippolito's research is supported in part by the DARPA KAIROS Program (contract FA8750-19-2-1004), the DARPA LwLL Program (contract FA8750-19-2-0201), and the IARPA BETTER Program (contract 2019-19051600004).  The views and conclusions contained herein are those of the authors and should not be interpreted as necessarily representing the official policies, either expressed or implied, of DARPA, IARPA, or the U.S. Government.

\section*{Contributions}
Each of the authors on this paper significantly contributed to the final results.
\begin{itemize}[leftmargin=*]
\item Katherine trained the models used in the paper, built and ran the eval and text generation pipelines, contributed significantly to writing, analysis, and project organization and management. 
\item Daphne ran the approximate matching data deduplication pipelines, extracted prompts and evaluation datasets, ran eval pipelines, and contributed significantly to planning, writing, and analysis.
\item Andrew wrote the code to perform deduplication with approximate matching, helped evaluate energy expenditure, and helped with analysis.
\item Chiyuan helped generate plots and contributed to project scoping, writing, and data analysis.
\item Chris offered mentorship and guidance throughout the project and contributed to writing.
\item Doug offered mentorship and guidance throughout the project and contributed to writing.
\item Nicholas wrote the suffix array implementation, ran all \Exact{} deduplication experiments, contributed significantly to planning, writing, and analysis, as well as scoping the project.
\end{itemize}

\bibliographystyle{acl_natbib}
\bibliography{main}
\appendix
\clearpage

\section{Further Details on \Approx{}}
\label{section:minhash_details}
For our MinHash based deduplication method, documents are first space tokenized, then each consecutive 5-gram is hashed using tabulation hashing.
The set of these hashes is the signature for the document.
For each element in a document's signature, the element is hashed using $k$ other hash functions.
The minimum hashed element for each of the $k$ hash functions is stored.
These minimum hashes are then partitioned into $r$ buckets, with $b$ hashes per bucket.
These $b$ hashes are augmented into a single value, then if two documents have the same value in at least one bucket, they'll be marked as a potential match.
The probability that two documents are considered a potential match is equal to
\begin{equation*}
\operatorname{Pr}(d_i, d_j | \operatorname{Jaccard}(d_i, d_j) = s_{i, j}) = 1 - (1 - s_{i, j}^b)^r
\end{equation*}
where $s_{i,j}$ is the Jaccard index between the two documents $i$ and $j$.
For document pairs that were identified as potential matches, we computed their actual Jaccard index, and if that was above 0.8, we computed their edit similarity.
Document pairs with edit similarity higher than 0.8 were identified as duplicates.
After some experimentation, we chose to use $b=20$, and $r=450$, so $k=9,000$, so as to make sure a collision at the desired Jaccard index threshold of 0.8 had a high probability of occurring.

\begin{figure*}[t]
    \centering
    \includegraphics[width=\linewidth]{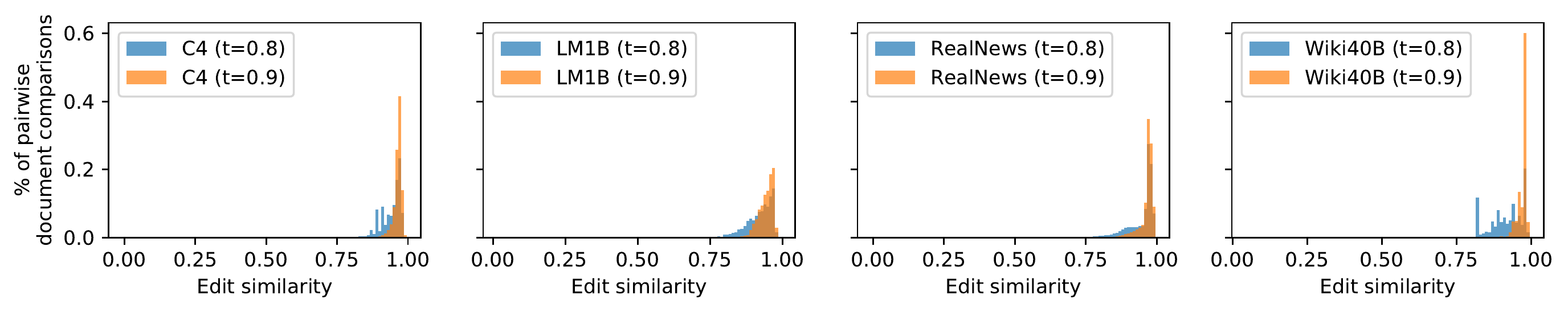}
    \includegraphics[width=\linewidth]{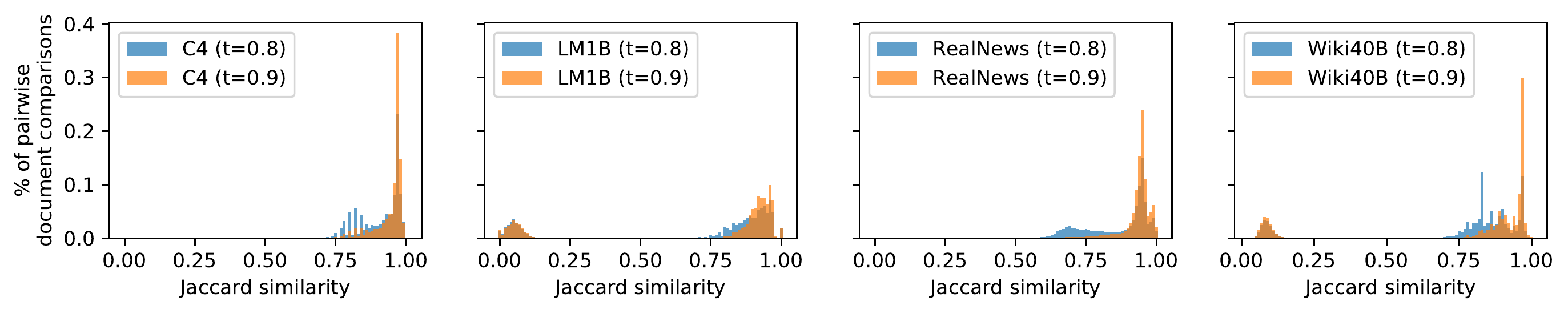}
    \caption{Histograms of document similarities.}
    \label{fig:neardup-hist}
\end{figure*}

We also tested an alternative configuration---filtering to document pairs with Jaccard index of at least 0.9 and edit similarity of at least 0.9.
In this case, we used $b=20$, $r=40$, and $k=800$.
Figure \ref{fig:neardup-hist} shows the histogram of Jaccard similarities and edit similarities for all document pairs which collided in min-hash space, for our chosen configuration (blue) and for the alternative configuration (orange).
This allows us verify if the threshold chosen has few comparisons around the chosen threshold, then we've likely captured the majority of actual near duplicates above that threshold. To verify that yourself, look at the left hand tails of the distributions. Since both 0.8 and 0.9 begin to vanish at the same point (in spite of the fact that the two thresholds are optimized for accuracy around different thresholds), we feel comfortable saying that we're capturing the majority of actual near duplicates. 

\paragraph{Computational Analysis}

Let $N$ be the number of documents and $T$ be the maximal number of tokens in a document. Edit similarity has a worst case complexity of $T^2$, so the worst case complexity is

\begin{equation*}
    O(N + b k^{2} T^{2} N) = O(N)
\end{equation*}

\noindent since $b$, $k$, and $T$ are all $\ll$ $N$. The left term is the complexity of grouping by the signatures, and the right represents the pathological worst case of all documents falling into the same $B$ buckets.

The highly distributed \Approx{} implementation we employed is one used for large-scale production tasks at Google.
On the English C4 dataset, the algorithm consumed approximately 41.5 kWh of energy.
Note that our choices of $k$ and $b$ were designed to produce very high recall, and with different parameters, the algorithm could be made much more energy efficient while producing similar results.

\section{Further Details on \Exact{}}
\label{sec:suffix-implementation}
\paragraph{Parallel linear time construction.}
We build a parallelized linear time suffix array algorithm.
As a building block, we make black-box use of the SA-IS algorithm for
constructing a suffix array in linear time \citet{nong2009linear,ko2003space}.
Unfortunately, this algorithm is not easily parallelized directly, so
we introduce a simple divide and conquer approach to parallelizing the array construction.

We build our implementation in Rust and extend an existing suffix array library\footnote{https://github.com/BurntSushi/suffix}
with three modification.
The first two are straightforward implementation differences:
we modify the code to allow datasets larger than $4$GB,
and we remove the requirement that strings parse as valid UTF-8 sequences in favor of raw byte sequences.
Our third change is more significant: we re-implement the algorithm
so that we can stream the suffix array itself off disk.

\paragraph{Parallel partial suffix array construction.}
Our divide and conquer suffix array construction algorithm starts by 
partitioning the dataset into $K$ different ``splits'' with SA-IS run
over independently on each split in parallel.
This algorithm still requires $O(N)$ work but runs in $O(N/K)$ wall-clock time.
This gives us $N$ separate suffix arrays $\mathcal{A}^i$.

Given two suffix arrays $A_1$ and $A_2$ for two sequences $S_1$ and $S_2$ it's not completely trivial to construct a single suffix array $A$ for $S = S_1 \mid\mid S_2$ 
because of the boundary conditions.
Instead, we don't build the data $S = S_1 \mid\mid S_2$ but rather let $S_1' = S_1 \mid\mid S_2[upto K]$ for some $K$ greater than the longest substring match.
Then we build the arrays on $S'_1$ and $S_2$.
To merge the arrays together we can remove the items from the first array after index $|S_1|$ and merge-sort insert them into the second.

\paragraph{Parallel merge of partial suffix arrays.}
We now merge these separate arrays together into a single suffix array $\mathcal{A}$,
Consider the simpler case of two partial suffix arrays $B$ and $C$ that we 
would like to merge together.
We can achieve this by letting $i=0$ index $B$ and $j=0$ index $C$.
Each iteration of the algorithm then pushes $B_i$ into $\mathcal{A}$ if
$S_{B_i..} < S_{C_i}$ and $C_i$ otherwise, repeating until $i=|B|-1$ and $j=|C|-1$.
To generalize to $K$ splits, we need only replace the single comparison above with
a min-heap requiring $O(\log{K}) \ll 10$ work on each iteration.

Observe that in the general case this algorithm is $O(N m \log(K))$ where $N$ is the length
of the dataset, $m$ is the average length of a prefix match, and $K$ is the number of splits.
It is therefore incorrect to call this algorithm linear time in the general case, for ours it is.
Because the length of the longest match is bounded above by the length of the longest
sequence, as long as the size of the dataset is independent of the length of the 
longest sequence in the dataset, this algorithm remains efficient.

Again, we can parallelize this operation among $L$ simultaneous jobs 
(in practice we set $K=L$ as the number of threads on our machine).
In the $K=2$ case, job $l$ processes $i \in [jN/L, (j+1)N/L]$, choosing
the bounds of $j$ by binary searching into $C$ so that $S_{B_{i}} < S_{C_{j}} < S_{B_{j+1}}$.
The case where $K>2$ is identical except that we repeat this over all $K$ partial suffix arrays.

\paragraph{Computational Analysis.}
We run our algorithm on a single VM on the cloud with $96$ cores and $768$GB of memory.
Our algorithm is efficient, for example processing the Wiki-40B training set ($3$ million
examples containing $4$GB of text) in $2.3$ minutes wall-clock time ($2.1$ CPU-hours of work).
The $350$GB C4 dataset takes under 12 hours (wall-clock) to build a suffix array; although we are still memory constrained and so this corresponds to $\sim 1000$ CPU-hours. 
Once the suffix array has been constructed, it takes under an hour to deduplicate the C4 dataset.

Note that this algorithm still requires that the dataset itself fits in memory
(so that we can efficiently index in arbitrary positions), but we do not need to fit the entire suffix array into memory.
This is fortunate since our suffix array requires an $8\times$ space overhead.
For example, the suffix array for the
$350$GB C4 is $1.5$TB.

Compared to the cost of training a language model on this dataset, the additional
work required to deduplicate the training dataset is negligible.

\begin{figure}
    \centering
    \includegraphics[scale=.8]{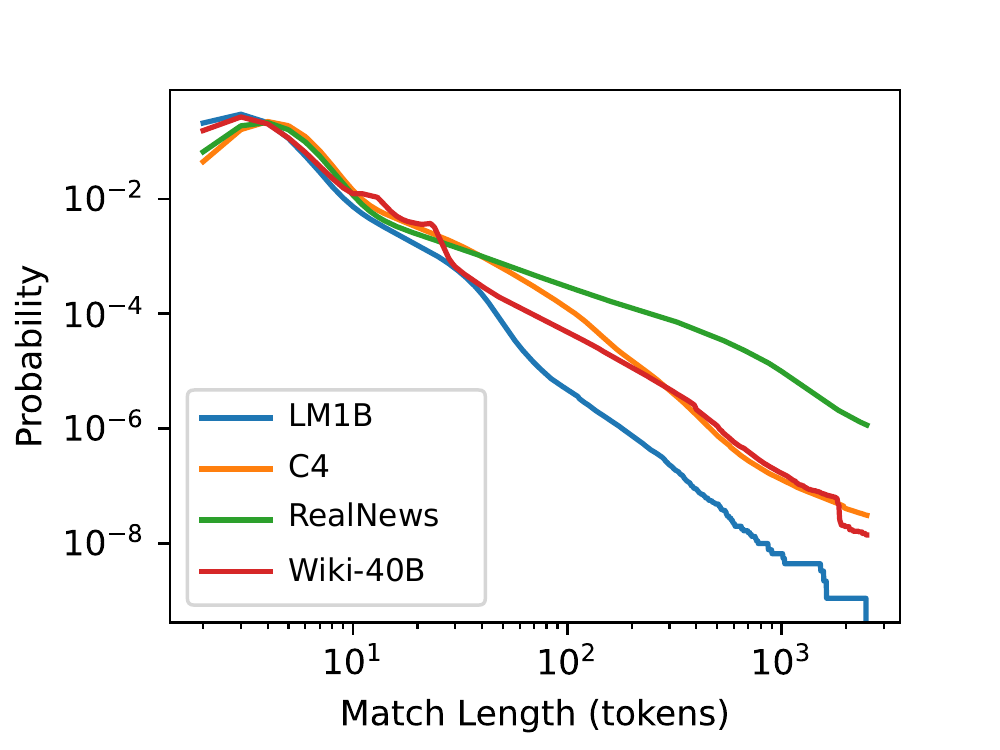}
    \caption{For each substring of length $k$,
    we plot the probability that there exists a second identical length-$k$ substring in the same train set.
    Matches with length under $10$ subword tokens are common, and account for $90\%$ of tokens.
    We choose a threshold of 50 for experiments.
    }
    \label{fig:suffix-match-len}
\end{figure}

\paragraph{Setting a threshold of duplicates.}
\label{section:exact_thresh}
An important question is how long must a substring match be before it is counted as a duplicate.
In Figure~\ref{fig:suffix-match-len}, we plot the frequency of substring matches within the four datasets we will
consider.
For each substring of length $k$, we compute the probability that there exists another sequence of length $k$ identical to this one; formally:
\[m(k) = \mathop{\text{Pr}}_{i \in [N]}\big[ \exists j \ne i : \mathcal{S}_{i..i+k} = \mathcal{S}_{j..j+k}\big].\]
We choose $50$ tokens as the threshold to be conservative:
the ``bend in the knee'' occurs at $10$ tokens, and manual inspection of
length-$25$ matches found no false positives.
We then doubled this value to have an exceptionally large margin for error.

\section{Further Details on Model Training}\label{sec:model-training-details}
Each model was trained for two epochs.
Since both C4-\Original{} and C4-\Exact{} contain approximately 365M examples, we performed 152K steps with a batch size of 4800 (or approximately 2 epochs). 
C4-\Approx{} contains approximately 350M examples, we performed 146K steps (or approximately 2 epochs).
On a 128-core TPU v3 pod slice, XL models trained on C4-\Original{} and C4-\Exact{} took approximately 131 hours (5.5 days) to train, while the XL model trained on C4-\Approx{} took approximately 126 hours to train.
Like T5, models were trained with the Adafactor optimizer \citep{shazeer2018adafactor}. A constant learning rate of 0.01 was used for the base models and 0.001 for the XL models.

The 1.5B parameter XL models had 24 layers, each with 32 attention heads. The model embedding size was 2,048, the feed forward layers had a hidden size of 5,120, and the key/value dimension size for the attention heads 64.
The 110M parameter base models had 12 layers, each with 12 attention heads.
The model embedding size was 768, the feed forward layers had a hidden size of 2,048, and the key/value dimension size for the attention heads 64.

\section{Energy Consumption} \label{sec:compute_energy}
\begin{table*}[t]
\centering
\small
\begin{tabular}{l|rrrrr}
\toprule
 & \multicolumn{1}{l}{T5 11B} & \multicolumn{1}{l}{\shortstack{XL-\Original\\ XL-\Exact}} & \multicolumn{1}{l}{XL-\Approx} & \multicolumn{1}{l}{\shortstack{Base-\Original\\Base-\Exact}} & \multicolumn{1}{l}{Total Inference} \\
 \midrule
TPU v3 cores & 512 & 128 & 128 & 64 & 64 \\
Training time (days) & 20 & 5.47 & 5.26 & 3 & 0.80 \\
TPU hrs & 245760 & 16804.70 & 16149.31 & 4608 & 1228.80 \\
Energy (MWh) & 85.70 & 5.86 & 5.63 & 1.61 & 0.43 \\
\bottomrule
\end{tabular}
\caption{Estimates of energy usage based on the data in \citet{patterson2021carbon}. The first column is \citet{patterson2021carbon}'s estimate of the T5 11B encoder-decoder model, which we based our own estimates on.
Inference includes all XL models. We generated 100,000 sequences from 3 models, with 5 prompts, and at 2 different checkpoints.). 
}
\label{tab:co2}
\end{table*}
We trained for approximately 131 hours or 5.5 days on a 128-core TPU v3.
The approximate deduplicated dataset is 3.9\% smaller than the original dataset and trains in 63 hours/epoch, saving us around 5 hours of compute time for the two epochs. 
The XL-\Original model was trained in North America 
where the XL-\Exact{} and XL-\Approx{} were trained in Taiwan. 
We used data from \citet{patterson2021carbon} to estimate amount of energy used in training these models
by computing the amount of $MWh$/hour/core and multiplying by our usage (see Table \ref{tab:co2} for how we computed these values).
For simplicity, we use estimates from Taiwainese datacenters as an estimate. 
We estimate training 2 epochs of XL-\Original{} and XL-\Exact{} uses $5.86 MWh$.
XL-\Approx{} is trained for fewer steps and we estimate uses $5.63 MWh$.
Training each base model was approximately 3 days on a 64-core TPU v3 pod slice which uses an estimated $1.61 MWh$. 

In addition to model training, evaluation and inference were performed on 64-core TPU v3 pod slices. Generating 100,000 sequences from the XL models takes approximately 0.64 hours. We generated 100,000 sequences for each of five types of prompts for two checkpoints of the model for a total of 1M sequences per model. This took approximately 19.2 hours. 
We estimate generating 3M sequences uses $0.43 MWh$. 

\begin{table*}[htbp]
  \small
  \centering
    \begin{tabular}{l|p{0.39\linewidth}|p{0.41\linewidth}}
    \toprule
    \multicolumn{1}{c|}{Dataset} & \multicolumn{1}{c|}{Example} & \multicolumn{1}{c}{Near-Duplicate Example} \\
    \midrule
    Wiki-40B & \pl{\textbackslash{}n\_START\_ARTICLE\_\textbackslash{}nHum Award for } {Most Impactful Character} \pl{\textbackslash{}n\_START\_SECTION\_\textbackslash{}nWinners and nominees\textbackslash{}n\_START\_PARAGRAPH\_\textbackslash{}nIn the list below, winners are listed first in the colored row, followed by the other nominees.} [...] &
    \pl{\textbackslash{}n\_START\_ARTICLE\_\textbackslash{}nHum Award for} {Best Actor in a Negative Role} \pl{\textbackslash{}n\_START\_SECTION\_\textbackslash{}nWinners and nominees\textbackslash{}n\_START\_PARAGRAPH\_\textbackslash{}nIn the list below, winners are listed first in the colored row, followed by the other nominees.} [...] \\
    \midrule
    LM1B  & \pl{I left for California in 1979 and tracked Cleveland 's changes on trips back to visit my sisters .} & \pl{I left for California in 1979} , \pl{and tracked Cleveland 's changes on trips back to visit my sisters .} \\
    \midrule
    RealNews & \pl{KUALA LUMPUR (Reuters) - Roads in Southeast Asia have been getting a little louder lately as motorcycle makers, an aspiring middle class and easy bank credit come together to breed a new genus of motorcyclists -- the big-bike rider. [...]} & A visitor looks at a Triumph motorcycle on display at the Indonesian International Motor Show in Jakarta September 19, 2014. REUTERS/Darren Whiteside\textbackslash{}n\pl{KUALA LUMPUR (Reuters) - Roads in Southeast Asia have been getting a little [...] big-bike rider.
    [...]} \\    
    \midrule
    C4    & \pl{Affordable and convenient holiday flights take off from your departure country,} "Canada"\pl{. From} May \pl{2019 to October 2019, Condor flights to your dream destination will be roughly} 6 \pl{a week! Book your} Halifax (YHZ) - Basel (BSL) \pl{flight now, and look forward to your} "Switzerland" \pl{destination!} &
    \pl{Affordable and convenient holiday flights take off from your departure country,} "USA"\pl{. From} April \pl{2019 to October 2019, Condor flights to your dream destination will be roughly} 7 \pl{a week! Book your} Maui Kahului (OGG) - Dubrovnik (DBV) \pl{flight now, and look forward to your} "Croatia" \pl{destination!} \\
    \bottomrule
    \end{tabular}%
  \caption{Qualitative examples of near-duplicates identified by \Approx{} from each dataset. The similarlity between documents is highlighted. Note the small interspersed differences that make exact duplicate matching less effective. Examples ending with ``[...]'' have been truncated for brevity.}
  \label{tab:qualitative_examples_appendix}%
\end{table*}%


\begin{table*}[htbp]
  \centering
    \small
    \begin{tabular}{p{.48\linewidth}|p{.48\linewidth}}
    \toprule
    \hl{Due to} high demand, \hl{we have} yet \hl{to critique this request. That said, we assure that the review will be produced in due time by our dilligent and} unwavering \hl{staff in a professional manner. This site is highly regarded amongst its peers in terms of} speed \hl{and reliability, so feel free to} check us out! &
    \hl{Due to} a heavy overflow, \hl{we have} not been able \hl{to critique this request. That said, we assure that the review will be produced in due time by our dilligent and} unshakable \hl{staff in a professional manner. This site is highly regarded amongst its peers in terms of} efficiency \hl{and reliability, so feel free to} visit! \\
    \midrule
    Need Pop Tacos parking? \hl{You can reserve parking near} Pop Tacos \hl{with SpotHero. Find low rates without parking coupons by booking a guaranteed spot online. Avoid circling, getting ticketed or running out to feed your meter. Search our parking map, compare parking rates and reserve a discounted parking spot today. Happy parking, and enjoy your meal at} Pop Tacos! & Il Sole parking. Reserve parking near Il Sole in NYC.\textbackslash{}n\hl{You can reserve parking near} Il Sole \hl{with SpotHero. Find low rates without parking coupons by booking a guaranteed spot online. Avoid circling, getting ticketed or running out to feed your meter. Search our parking map, compare parking rates and reserve a discounted parking spot today. Happy parking, and enjoy your meal at} Il Sole! \\
    \midrule
    \hl{This item was available on} Vinyl 7" \hl{but is now sold out on all formats, sorry. Take a look at what else we have in by} Jumbo\hl{, check out some related artists, head over to our new releases or knock yourself out reading our latest music news \& album reviews.}\textbackslash{}n2nd single edn of 550. &
    \hl{This item was available on} CD \hl{but is now sold out on all formats, sorry. Take a look at what else we have in by} Sirconical, Misty Dixon, Various\hl{, check out some related artists, head over to our new releases or knock yourself out reading our latest music news \& album reviews.}\textbackslash{}nTwisted Nerve comp mini album. \\
    \midrule
    \hl{Here is all the information you need about} "No One Killed Jessica" on American Netflix. \hl{Details include the date it was added to} Netflix in the USA\hl{, any known expiry dates and new episodes/seasons, the ratings and cast etc. So scroll down for more information or share the link on social media to let your friends know what you're watching.} &
    \hl{Here is all the information you need about} "A Land Imagined" on Netflix in the UK. \hl{Details include the date it was added to} UK Netflix\hl{, any known expiry dates and new episodes/seasons, the ratings and cast etc. So scroll down for more information or share the link on social media to let your friends know what you're watching.} \\
    \midrule
    8 + 8 = \hl{Solve this simple math problem and enter the result. E.g. for 1+3, enter 4.}
    & Math question * 7 + 1 = \hl{Solve this simple math problem and enter the result. E.g. for 1+3, enter 4.} \\
    \midrule
    Long Island College Hospital \hl{is committed to providing outstanding patient care in the} Brooklyn, NY area\hl{, but before you commit to} Long Island College Hospital for a Endometrial Ablation \hl{make sure you compare and shop other medical facilities. It may save you hundreds (in some cases thousands) of dollars. View a} Endometrial Ablation \hl{cost comparison for} Brooklyn and \hl{Request a Free Quote before you make a decision.} &
    Morristown Memorial Hospital \hl{is committed to providing outstanding patient care in the} Morristown, NJ area\hl{, but before you commit to} Morristown Memorial Hospital for a Breast Ultrasound \hl{make sure you compare and shop other medical facilities. It may save you hundreds (in some cases thousands) of dollars. View a} Breast Ultrasound \hl{cost comparison for} Morristown and \hl{Request a Free Quote before you make a decision.} \\
    \bottomrule
    \end{tabular}%
  \caption{Several examples of pairs of documents in C4 that were found by the Approximate Matching algorithm and identified as having edit similarity of almost exactly 0.8. Pairs of documents less similar than 0.8 were not identified as duplicates. For readability, matching subsequences have been highlighted.}
  \label{tab:qualitative_near_boundary}%
\end{table*}%

\begin{table*}[h]
  \centering
  \small
    \begin{tabular}{p{0.8\linewidth}|r}
    \toprule
    \multicolumn{1}{c|}{\textbf{Text}}  & \multicolumn{1}{c}{\textbf{Freq in C4}} \\
    \midrule
    HD wallpaper. This wallpaper was upload at April 19, 2019 upload by admin in.You can download it in your computer by clicking resolution image in Download by size:. Don't forget to rate and comment if you interest with this wallpaper. &            40,340  \\
    \hline
    to the address posted below. Include our failure information form,a packing slip with your Company name, contact person, and Email address or phone number. Upon receipt of your repair, we\textbackslash{}'ll inspect it and then contact you with a quote or evaluation notice. Normal turn around for repair is 5 to 7 business days, with "Rush Repair" available. &              5,900  \\
    \hline
    is a great place to begin your search. Whether you are a first-time home buyer or you are already familiar with the home buying process, you can be assured that you have the best tools and the perfect agent available to help with your &              5,358  \\
    \hline
    pics at these awesome group starting P letter. Desktop wallpapers were first introduced way back in the 1980s and have gained immense popularity since then. It is possible to come across more than 80 million sites on the web offering some sort of wallpaper. &                 848  \\
    \hline
    flowers will let them know you're thinking of them and wishing them well. Cheerful yellow flowers bring their own sunshine and will get right to work on lifting spirits, and a colorful vase will bring loads of smiles to friends and visitors! Get Well flower arrangements from &                 479  \\
    \hline
    our premier 24 hour emergency* plumbing and heating solutions. We realise that when your heating fails or pipes and drains leak it can cause havoc with your routine and even cause damage to your property. When a plumbing problem occurs that requires an immediate response we provide qualified local plumbers throughout &                   56  \\
    \hline
     is to remove all images that violate copyrights. Please contact us to request that images be removed or to assign proper credit. The images displayed on this site may be used for Free or educational purposes only. If you would like to use any of the images displayed on this site for any other purpose, please obtain permission from the owner. www. &                   48  \\
     \hline
     list of fishing locations, providing interactive maps that show each location's GPS coordinates, nearby facilities (like restaurants, gas stations, marinas and fishing shops), their current and forecasted weather and, if available, their water conditions.\textbackslash{}nFind any of the 8 &                     5  \\
     \hline
    . Dyer, Ph.D., is an internationally renowned author and speaker in the field of self-development. He's the author of 30 books, has created many audio programs and videos, and has appeared on thousands of television and radio shows. &                     5  \\
    \bottomrule
    \end{tabular}%
\caption{A selection of substrings identified by \Exact{} as being in C4 multiple times. The number of times this exact substring occurs in C4 is also given.}
  \label{tab:exact_substr_examples}%
\end{table*}%

\begin{table*}[h]
  \centering
  \small
    \begin{tabular}{p{0.8\linewidth}|r}
    \toprule
    \multicolumn{1}{c|}{\textbf{Generated Text}}  & \multicolumn{1}{c}{\textbf{Freq in C4}} \\
    \midrule
    , you'll need to be knowledgeable to make the very best decisions. We will make sure you know what can be expected. We take the surprises from the picture by giving accurate and thorough information. You can start by talking about your task with our client service staff when\newline{} you dial 888-353-1299. We'll address all of your questions and arrange the initial meeting. We work closely with you through the whole project, and our team can show up promptly and prepared. & 5,497 \\
    \hline
    then Waterside Lodge are well equipped for the task. Our fully equipped family sized lodges offer a comfortable luxurious stay for a fantastic price, giving you beautiful views of the lakes and the surrounding countryside. Offering luxurious self-catering holidays in our fully featured Scandinavian holiday lodges. Perfectly located to explore the beaches, coastline. All of our lodges are sized for 6 people and are furnished to the highest standards to ensure you have a stay like no other. At Waterside Lodge the stay itself is only half of the package, Waterside lodge is situated closely to the Heritage Coast which makes our lodges the perfect stay for anyone wanting to get away and have a relaxing countryside break from the city. Whilst you stay with us be sure to take advantage of all the activities Waterside Lodge has to offer. Such as the use of our on-site fishing lakes for the keen fisherman, free internet access, outside relaxation areas, comfortable lounges and much more. & 571 \\
    \hline
    you are only looking to find rent to own homes in your city or are open to exploring all kinds of rent to own home listings, our database does it all. One of the best aspects of iRentToOwn.com is that, besides options to rent to buy a house, it has numerous other categories of home sale options. These include bank foreclosure homes, pre-foreclosure homes, short sales, HUD/government foreclosures, auction homes and owner-financing/FSBO (For Sale By Owner) homes. With help from the convenient search features offered by our site, shoppers are able to find their ideal lease to own home, real estate company, and more in South & 51 \\
    \hline
    , IL employs journeyman as licensed to work by themselves, without direct supervision, installing wiring, outlets and fixtures. Our journeyman also does service work, troubleshooting when a breaker fails or a light stops working. Our journeyman does not offer permits that must be issued by our master. Our journeyman follows our master's plans and directions. Our journeyman's responsibilities will vary based on the work that needs to be done. Our journeymen are skilled with residential, commercial and industrial installations and repairs.ust work from six years as an apprentice, under direct supervision of our master, and pass a journeyman test. This person also must have some classroom education on the National Electrical Code and fundamental electricity in a technical school a program affiliated with the National Joint Apprenticeship Training Council. Journeyman training combines hands-on work with education on basic electricity. & 6 \\
    \hline
    combustion process of a petrol engine is never perfect. Dangerous gases, such as nitrogen oxide, carbon monoxide and hydrocarbons will arise and it is the job of the catalytic converter to reduce these to safer emissions. These cat converters can fail by becoming clogged, or if the engine has bad exhaust valves or the plugs fail, causing unburned fuel to overheat the converter. Mettam's Mufflers can resolve these issues with your Karr & 5 \\
    \hline
    ,ANDREW Find the ancestral town: Many a researcher is stuck behind records that say, BIRTHPLACE: IRELAND without saying where in Ireland, or whatever other country. Remember that your immigrant ancestor's siblings probably were born in the same ancestral town, so check all o\newline{}f their records, too. Around 1900, the Roman Catholic churches reported marriages to the churches where the persons were baptised, and before the wedding, they would require a baptismal certificate from that church, without marriage notations, to make sure that the persons were no\newline{}t already married, ordained, or whatever, and were free to marry. Do check the Catholic records especially for ex loco and the home town. If your ancestor's sister had a daughter who generated a marriage or death record saying, MOTHER'S BIRTHPLACE: and the exact town, then y\newline{}ou know where to start searching for records that will confirm it is your ancestor's home town. BEWARE: Just because you find a family with the same names does not mean they are the same family, as they could very well be an unrelated family from a different town in the same an\newline{}cestral country. The webmaster has learned this. One clue was that one family was still having babies in Potenza city, Italy while the other was having babies in Colorado, U.S.A. & 2 \\
    \hline
    will not want to search for Power Washing companies in Wyoming on an extensive basis. The service personnel will be at your doorsteps through online or phone booking. The power wash solutions offered by us are matchless and you can compare with others in Winfield, IL. The power wash services offered by us are very economical. Gutter brightener will be applied which will be followed by cleaning through double scrub. The cleaning will be done by using a soft bristle brush. The bond and contaminants will be released in an effortless manner. & 1 \\
    \hline
    Z3 Plus are valid in all major cities of India like Delhi, Gurgaon, Noida, Mumbai, Chennai, Bangalore, Hyderabad, Kolkata, Pune, Ahmedabad, Coimbatore, Lucknow, Trichy, Madurai, Trivandrum, Mysore, Jaipur, Chandigarh, Pondicherry, Bhopal, Patna, Bhubaneswar, Amritsar, Cochin, \newline{}Allahabad, Srinagar, New Delhi, Surat, Ludhiana, Navi Mumbai, Ghaziabad, Bengaluru, Indore, Nagpur, Thane, Agra, Meerut, Ranchi. The delivery feasibility and charges may be varying, hence for them please check with the particular seller or store. & 1 \\
    \bottomrule
    \end{tabular}%
\caption{A selection of substrings generated by XL-\Original{} with no prompting (and top-$k$ with $k$=50) that were identified by \Exact{} as being in C4 multiple times. The number of times each substring was found in C4 is given. We observe that most memorized generations tend to be from advertisements.}
  \label{tab:approx_gen_noprompt_examples}%
\end{table*}%

\begin{table*}[ttbp]
  \centering
  \small
    \begin{tabular}{lrr||lrr}
    \toprule
    \multicolumn{1}{c}{RealNews Url} & \multicolumn{1}{c}{\# Total} & \multicolumn{1}{c}{Frac Dups} & \multicolumn{1}{c}{C4 Url} & \multicolumn{1}{c}{\# Total} & \multicolumn{1}{c}{Frac Dups} \\
    \hline
    medicalnewstoday.com. & 12    & 1.00  & hairtechkearney.com & 4883  & 1 \\
    dodbuzz.com & 301   & 0.99  & keywordsking.com & 1786  & 1 \\
    undertheradar.military.com & 187   & 0.97  & sydneysitalianfruitshops.online & 1178  & 1 \\
    q.usatoday.com & 33    & 0.94  & moewiki.usamimi.info & 1001  & 1 \\
    ad-test.thirdage.com & 354   & 0.94  & swarovskijewelryoutlet.org & 984   & 1 \\
    amp.nymag.com & 15    & 0.93  & forzadurto.org & 980   & 1 \\
    citizenwire.com & 1022  & 0.93  & producerati.com & 971   & 1 \\
    paycheck-chronicles.military.com & 363   & 0.92  & sourceryforge.org & 908   & 1 \\
    product-reviews.net & 73403 & 0.92  & heavenz-kitchen.com & 876   & 1 \\
    kitup.military.com & 196   & 0.92  & little-eclipse.com & 822   & 1 \\
    gcaptain.com & 33903 & 0.92  & walops.com & 819   & 1 \\
    dev.screenrant.com & 70    & 0.91  & 16thstlaunderland.com & 713   & 1 \\
    live.swissinfo.ch & 66    & 0.91  & theroyalstarinfo.com & 696   & 1 \\
    news.theepochtimes.com & 82    & 0.87  & code4kt.com & 684   & 1 \\
    opinion.toledoblade.com & 986   & 0.87  & nflfalconsjerseys.us & 682   & 1 \\
    cdn.moneytalksnews.com & 121   & 0.86  & quiltingbeeshop.com & 676   & 1 \\
    amp.fox23.com & 14    & 0.86  & ulifeinsurancemiami.com & 675   & 1 \\
    sales.rollingstone.com & 20    & 0.85  & wowkeyword.com & 673   & 1 \\
    ftp.screenrant.com & 20    & 0.85  & taspetro.com & 671   & 1 \\
    \bottomrule
    \end{tabular}%
  \caption{On the left, we show the URLs that had the greatest proportion of examples marked as near-duplicates by \Approx (filtered to URLs which occurred at least 10 times). On the right, we show the 20 most frequent URLs in C4 for which all examples were marked as near-duplicates by \Approx.}
  \label{tab:urls}%
\end{table*}%
\section{More Results}
\paragraph{Qualitative Examples.}
Table \ref{tab:qualitative_near_boundary} shows several examples of pairs of documents in C4 whose edit distance is close to our chosen edit similarity threshold of 0.8.
Table \ref{tab:exact_substr_examples} shows substrings which were identified by \Exact{} as being in C4 more than once.
Table \ref{tab:approx_gen_noprompt_examples} shows several examples of unprompted generations which were identified as memorized are shown.

\begin{figure}
    \centering
    \includegraphics[width=\linewidth]{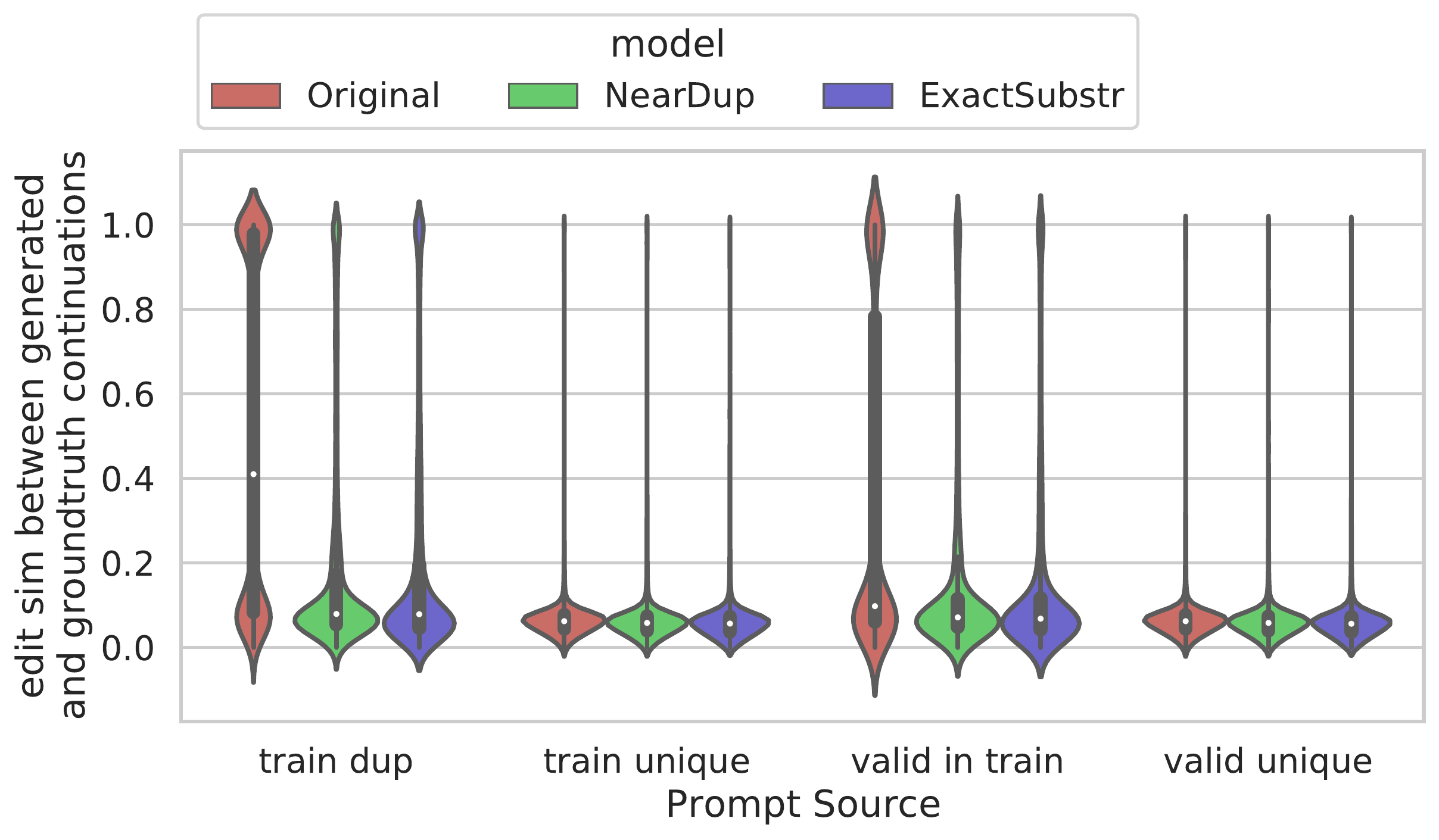}
    \caption{Memorized continuations distribution}
    \label{fig:mem-cont-dist}
\end{figure}

\paragraph{Distribution of memorization.}
Figure \ref{fig:mem-cont-dist} shows the distribution in memorization amount over all generated sequences when using four types of prompting: train example with duplicates in train, train examples without any duplicates, validation examples with duplicates in train, and validation examples without any duplicates.

\paragraph{URLs with many duplicates.}
Table \ref{tab:urls} shows the URLs had the largest proportion of examples identified by \Approx{} as near-duplicates. 
For C4, these tend to be websites that sell many similar products and thus have a large amount of templated text.
For RealNews, content aggregators seem especially common.

\begin{figure}
    \centering
    \includegraphics[width=\linewidth]{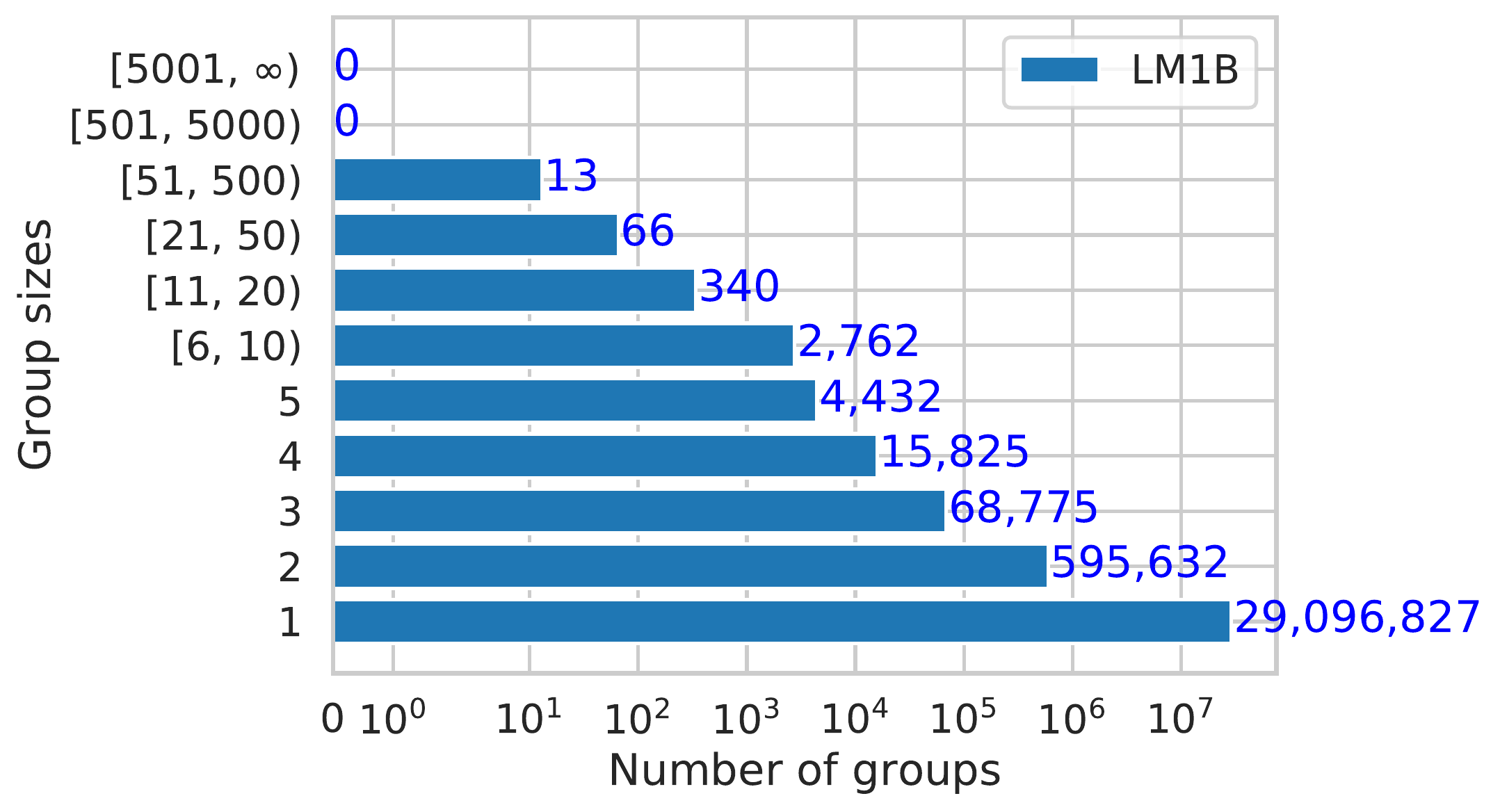}
    \includegraphics[width=\linewidth]{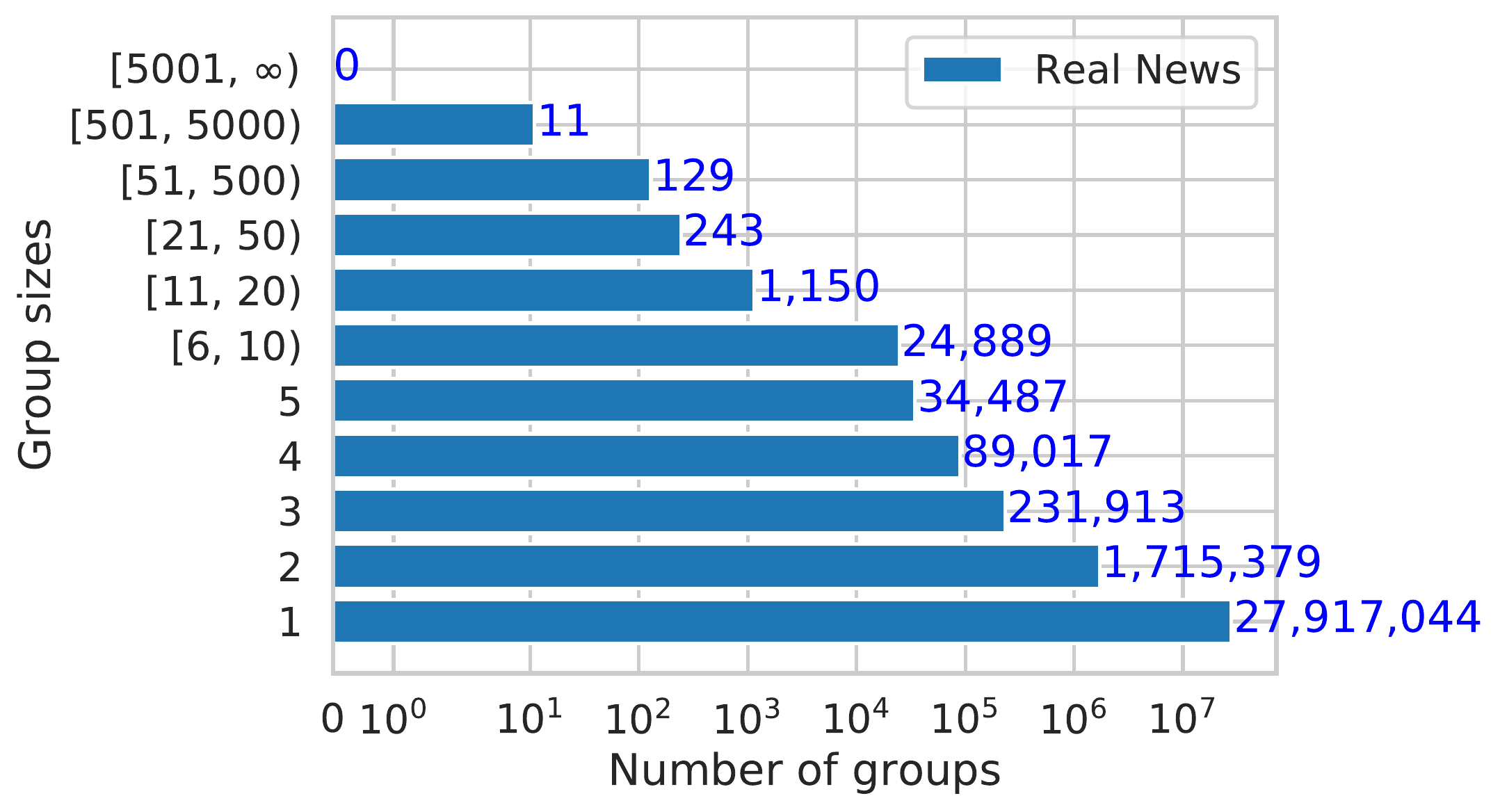}
    \includegraphics[width=\linewidth]{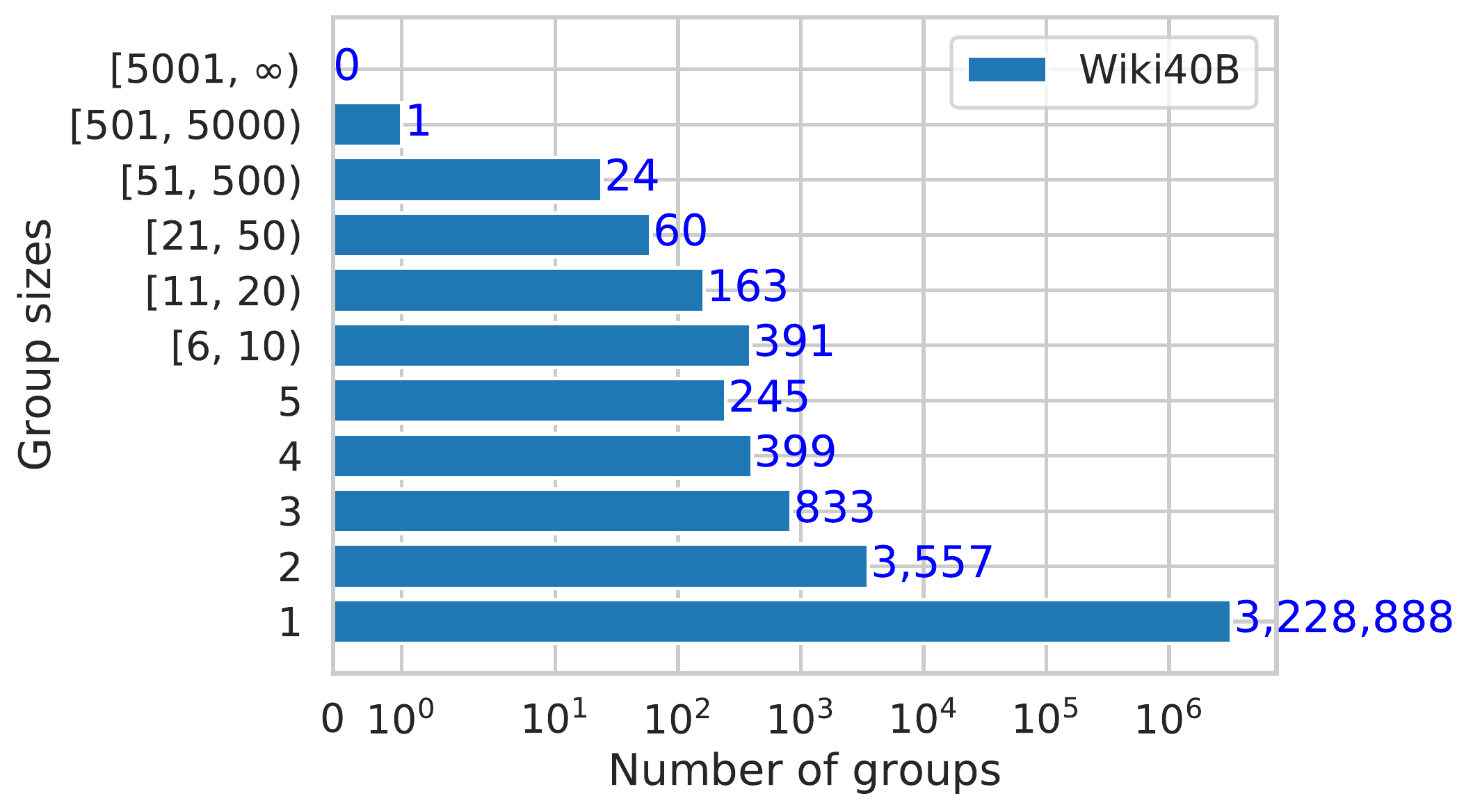}
    \caption{The distribution of near-duplicate cluster sizes from running \Approx{} on each dataset.}
    \label{fig:nd3-cluster-hist}
\end{figure}

\paragraph{\Approx{} cluster sizes.}
Figure \ref{fig:nd3-cluster-hist} shows the distribution of cluster sizes from running \Approx{} on RealNews, LM1B, and Wiki-40B (results for C4 are in Figure \ref{fig:nd3-cluster-hist-c4} the main paper).

\paragraph{Dataset Sizes}
Table \ref{tab:dataset_sizes} gives the size in BPE tokens and in examples of each dataset before and after deduplication.
Because most datasets were already deduplicated of exact matches during their creation, \Exact deduplication does not actually remove any examples.

\begin{figure*}[t]
    \centering
    \begin{overpic}[width=.5\linewidth]{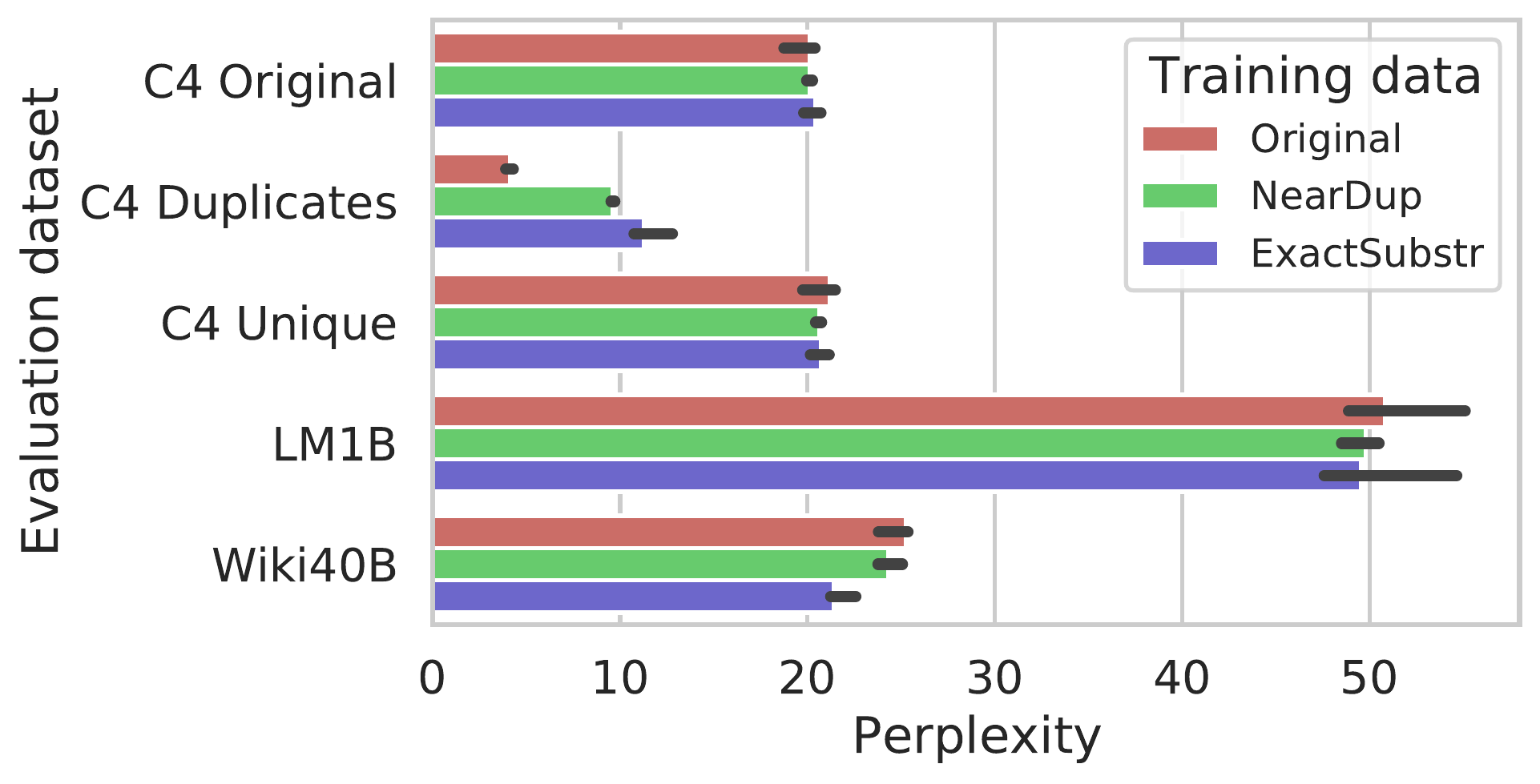}
    \put(1,1){\small\textbf{(a)} Base model}
    \end{overpic}\vskip5pt
    \begin{overpic}[width=.5\linewidth]{figs/eval-xl-ppl_withLM1B.pdf}
    \put(1,1){\small\textbf{(b)} XL model}
    \end{overpic}
    \caption{Impact of deduplicating the training set on validation perplexity. In \textbf{(a)}, we plot the results from T5 base (110M parameters) across three training runs with different random initializations. The black bar represent the lowest perplexity to the highest perplexity, and the colored bar the median perplexity. 
    In \textbf{(b)}, we plot the results from T5 XL (1.5B parameters).}
    \label{fig:eval-ppl-with-lm1b}
\end{figure*}

\paragraph{Perplexity on LM1B.}
Figure~\ref{fig:eval-ppl-with-lm1b} is the same as Figure~\ref{fig:eval-ppl} of the main paper, except with perplexity on LM1B included.
LM1B was omitted from the main paper's figure in order to improve readability. 

\begin{table*}[ht]
    \centering
    \small
\begin{tabular}{l|rrrrrr}
\toprule
  \multicolumn{1}{l}{Training Dataset:}  & \multicolumn{2}{c}{C4-\Original} & \multicolumn{2}{c}{C4-\Approx{}} & \multicolumn{2}{c}{C4-\Exact{}} \\
 \multicolumn{1}{l}{Epoch:} & \multicolumn{1}{r}{1} & \multicolumn{1}{c}{2} & \multicolumn{1}{c}{1} & \multicolumn{1}{c}{2} & \multicolumn{1}{c}{1} & \multicolumn{1}{c}{2} \\
 \midrule
No prompt & 1.93\% & 1.57\% & 0.19\% & 0.26\% & 0.14\% & 0.17\% \\
Duplicate Train Prompts & 35.88\% & 34.34\% & 3.34\% & 3.15\% & 5.71\% & 4.67\% \\
Unique Train Prompt & 0.42\% & 0.41\% & 0.42\% & 0.41\% & 0.22\% & 0.23\% \\
Duplicate Test Prompt & 16.27\% & 15.32\% & 1.61\% & 1.52\% & 0.34\% & 0.25\% \\
Unique Test Prompt & 0.25\% & 0.22\% & 0.21\% & 0.23\% & 0.03\% & 0.08\% \\

\bottomrule
\end{tabular}
\caption{Percentage of tokens in 100k generations that were part of memorized substring according to \Exact.
Models trained with approximate or exact deduplication have 
$10\times$ less memorization than the model trained on the original (non-deduplicated) dataset.}
\end{table*}
\begin{table*}[t]
  \centering
  \small
    \begin{tabular}{l|rrr|rrr}
    \toprule
          & \multicolumn{3}{c|}{Final train set size in tokens} & \multicolumn{3}{c}{Final train set size in examples} \\
          & \Original & \Approx & \Exact & \Original & \Approx & \Exact \\
          \hline
    C4    & 177.3B & 173.7B & 165.4B & 364.87M & 350.48M & 350.48M \\
    Real News & 24.7B & 22.4B & 20.1B  & 31.16M & 28.39M & 28.39M \\
    LM1B  & 1.0B & 0.94B  & 0.90B  & 30.30M & 29.87M & 30.16M \\
    Wiki40B & 2.25B & 2.24B & 2.19B  & 2.93M & 2.91M & 2.93M \\
    \bottomrule
    \end{tabular}%
  \caption{Each row shows the size in tokens (according to our 50k BPE vocab) and in examples of a train set in its original form, with \Approx{} deduplication, and with \Exact{} deduplication.}
  \label{tab:dataset_sizes}%
\end{table*}%

\end{document}